\documentclass[10pt]{article} 
\usepackage[preprint]{tmlr}

\usepackage{amsmath,amsfonts,bm}

\def\eqref#1{equation~\ref{#1}}

\def\1{\bm{1}}

\DeclareMathAlphabet{\mathsfit}{\encodingdefault}{\sfdefault}{m}{sl}
\SetMathAlphabet{\mathsfit}{bold}{\encodingdefault}{\sfdefault}{bx}{n}

\usepackage{hyperref}
\usepackage{url}
\usepackage{graphicx}
\usepackage{multirow}
\usepackage{booktabs}       

\usepackage[capitalize, noabbrev]{cleveref}
\usepackage[section]{placeins}
\usepackage{wrapfig}
\usepackage{caption}
\usepackage{subcaption}
\usepackage{colortbl}

\newcommand{\hpi}{{\hat{\pi}}}
\newcommand{\hs}{{\hat{s}}}
\newcommand{\hP}{{\hat{P}}}
\newcommand{\hV}{{\hat{V}}}

\title{Offline Policy Comparison with Confidence: Benchmarks and Baselines}

\author{
    \name Anurag Koul \email koula@oregonstate.edu \\
    \addr School of EECS\\
    Oregon State University
    \AND
    \name Mariano Phielipp \email mariano.j.phielipp@intel.com \\
    \addr Intel Labs
    \AND
    \name Alan Fern \email afern@oregonstate.edu \\
    \addr School of EECS \\ 
    Oregon State University
}

\def\month{MM}  
\def\year{YYYY} 
\def\openreview{\url{https://openreview.net/forum?id=XXXX}} 

\begin{document}

\maketitle

\begin{abstract}
    Decision makers often wish to use offline historical data to compare sequential-action policies at various world states. Importantly, computational tools should produce confidence values for such offline policy comparison (OPC) to account for statistical variance and limited data coverage. Nevertheless, there is little work that directly evaluates the quality of confidence values for OPC. In this work, we address this issue by creating benchmarks for OPC with Confidence (OPCC), derived by adding sets of policy comparison queries to datasets from offline reinforcement learning. In addition, we present an empirical evaluation of the \emph{risk versus coverage} trade-off for a class of model-based baselines. In particular, the baselines learn ensembles of dynamics models, which are used in various ways to produce simulations for answering queries with confidence values. While our results suggest advantages for certain baseline variations, there appears to be significant room for improvement in future work.
\end{abstract}

\section{Introduction}
\label{sec:introduction}
Given historical data from a dynamic environment, how well can we make predictions about future trajectories while also quantifying the uncertainty of those predictions? Our main goal is to drive research toward a positive answer by encouraging work on a specific prediction problem, \emph{offline policy comparison with confidence (OPCC)}. Toward this goal we contribute OPCC benchmarks with metrics that directly relate to uncertainty quantification along with a baseline pilot evaluation.
The \emph{benchmarks} \footnote{ Benchmarks: \url{https://github.com/koulanurag/opcc}} and \emph{baselines}\footnote{ Baselines:  \url{https://github.com/koulanurag/opcc-baselines}} are made publicly available.

OPCC involves using historical data to answer queries that each ask for: 1) a prediction of which of two policies is better for an initial state and horizon, where the policies, state, and horizon can be arbitrarily specified, and 2) a confidence value for the prediction. While here we use OPCC for benchmarking uncertainty quantification, it also has utility for both decision support and policy optimization. For decision support, a farm manager may want a prediction for which of two irrigation policies will best match season-level crop goals. A careful farm manager, however, would only take the prediction seriously if it comes with a meaningful measure of confidence. For policy optimization, we may want to search through policy variations to identify variations that confidently improve over others in light of historical data.

Offline reinforcement learning (ORL) \citep{levine2020offline}, both for policy evaluation and optimization, offers a number of techniques relevant to decision support and OPCC in particular. One of the key ORL challenges is dealing with uncertainty due to statistical variance and limited coverage of historical data. This recognition has led to rapid progress in ORL, yielding different approaches for addressing uncertainty, e.g. pessimism in the face of uncertainty \citep{kumar2020conservative, buckman2020importance, jin2021pessimism, shrestha2021} or regularizing policy learning toward the historical data \citep{fujimoto2021minimalist,kostrikov2021offline,kumar2019stabilizing,peng2019advantage}. However, there has been very little work on directly evaluating the uncertainty quantification capabilities embedded in these approaches. Rather, overall ORL performance is typically evaluated, which can be affected by many algorithmic choices that are not directly related to uncertainty quantification. A major motivation for our work is to better measure and understand the underlying uncertainty quantification embedded in popular ORL approaches.

Prior work has studied non-sequential prediction (e.g. image classification) with an abstention (or rejection) option \citep{el-yaniv2010,geifman2017,geifman2019,hendrickx2021,xin2021, condessa2017performance}. Typically, these methods produce confidence values for predictions and abstain based on a confidence threshold. Ideally, if the confidence values strongly relate to prediction uncertainty, then abstentions will be biased toward the erroneous predictions. In order to directly evaluate the quality of uncertainty quantification, this line of work commonly reports measures of risk-coverage curves (RCCs) such as area under the curve (AUC) and reverse-pair proportion (RPP). To the best of our knowledge, analogous benchmarks and evaluations have not yet been established for sequential decision making. Our focus on establishing benchmarks and metrics for OPCC aims at partially filling this gap.

\textbf{Contribution. }The first contribution of this paper is to develop benchmarks for OPCC derived from existing ORL benchmarks and to suggest metrics for the quality of uncertainty quantification. Each benchmark includes: 1) a set of trajectory data $D$ collected in an environment via different types of data collection policies, and 2) a set of queries $Q$, where each query asks which of two provided policies has a larger expected reward with respect to a specified horizon and initial states. Note that our OPCC benchmarks are related to recent benchmarks for offline policy evaluation (OPE) \citep{fu2021benchmarks}, which includes a policy ranking task similar to OPCC. That work, however, does not propose evaluation metrics and protocols for measuring uncertainty quantification over policy rankings. Further, our query sets $Q$ span a much broader range of initial states than existing benchmarks, which is critical for understanding how uncertainty quantification varies across the wider state space as it relates to the trajectory data $D$. The benchmarks and baselines are publicly available with the intention of supporting community expansion over time. 

Our second contribution is to present a pilot empirical evaluation of OPCC for a class of approaches that use ensembles as the mechanism to capture uncertainty, which is one of the prevalent approaches on ORL. This class uses learned ensembles of dynamics and reward models to produce Monte-Carlo simulations of each policy, which can then be compared in various ways to produce a prediction and confidence value. Our results for different variations of this class provide evidence that some variations may improve aspects of uncertainty quantification. However, overall, we did not observe sizeable and consistent improvements from most of the variations we considered. This suggests that there is significant room for future work aimed at consistent improvement for one or more of the uncertainty-quantification metrics.

\section{Background}

We formulate our work in the framework of Markov Decision Processes (MDPs), for which we assume basic familiarity \citep{puterman2014}. An MDP is a tuple $M= \{S,A, P, R\}$, where $S$ is the state space, $A$ is the action space, and $P(s'|s,a)$ is the first-order Markovian transition function that gives the probability of transitioning to state $s'$ given that action $a$ is taken in state $s$. Finally, $R(s, a)$ is potentially a stochastic reward function, which returns the reward for taking action $a$ in state $s$.

In this work, we focus on decision problems with a finite horizon $h$, where action selection can depend on the time step. A non-stationary policy $\pi(s,t)$ is a possibly stochastic function that returns an action for the specified state $s$ and time step $t\in \{0,\ldots,h-1\}$. Given a horizon $h$ and discount factor $\gamma \in [0,1)$, the value of a policy $\pi$ at state $s$, denoted $V^{\pi}(s,h)$, is the expected cumulative discounted future reward over the horizon:
\begin{small}
    \begin{equation*}
        V^{\pi}(s,h) = \mathbb{E}\left.\left[\sum_{t=0}^{h-1} \gamma^{t} R\left(S_{t}, A_{t}\right) \right| S_0 = s, A_t = \pi(S_t,t)\right]
    \end{equation*}
\end{small}
\noindent where $S_t$ and $A_t$ are the state and action random variables at time $t$. It is important to note that we gain considerable flexibility by allowing for non-stationary policies. For example, $\pi$ could be an open-loop policy or even a fixed sequence of actions, which are commonly used in the context of model-predictive control \citep{richards2005robust}. Further, we can implicitly represent the action value function $Q^{\pi}(s,a,h)$ for a policy $\pi$ by defining a new non-stationary policy $\pi'$ that takes action $a$ at $t=0$ and then follows $\pi$ thereafter, which yields $V^{\pi'}(s,h) = Q^{\pi}(s,a,h)$. For this reason, we will focus exclusively on comparisons in terms of state-value functions without loss of generality.

\section{Offline Policy Comparison with Confidence}

In this section, we first introduce the concept of policy comparison queries, which are then used to define the OPCC learning problem. Finally, we discuss the OPCC evaluation metrics used in our evaluations.

\subsection{Policy Comparison Queries}

We consider the fundamental decision problem of predicting the relative future performance of two policies, which we formalize via \emph{policy comparison queries (PCQs)}. A PCQ is a tuple $q=(s, \pi, \hs, \hpi, h)$, where $s$ and $\hs$ are arbitrary starting states, $\pi$ and $\hpi$ are policies, and $h$ is a horizon. The answer to a PCQ is the truth value of $V^{\pi}(s,h) < V^{\hpi}(\hs,h)$. That is, a PCQ asks whether the $h$-horizon value of $\hpi$ is greater than $\pi$ when started in $\hs$ and $s$ respectively.

As motivated in Section \ref{sec:introduction}, PCQs are useful for both human-decision support and automated policy optimization. For example, if a farm manager wants information about which of two irrigation policies, $\pi$ and $\hpi$, will result in the best future crop yield given the environment state $s$, then the corresponding PCQ would be $(s, \pi, s, \hpi, h)$. Alternatively, the manager may be interested in whether a policy $\pi$ is better suited to an environmental state $s$ or $\hs$, which is captured by the PCQ $(s, \pi, \hs, \pi, h)$. In addition, PCQs can be used as the basis for the classic policy improvement step of policy iteration \citep{puterman2014}. In particular, we can improve over policy $\pi$ at state $s$ by identifying an action $a'$ with higher action value than chosen by $\pi$. The corresponding PCQ for testing $a'$ is $(s, \pi, s, \pi', h)$, where $\pi'$ is the non-stationary policy that first takes action $a'$ and then follows $\pi$. 

In practice, PCQs within an application domain need not be restricted to comparing policies via a single reward function. Rather there are often multiple quantities of interest to users. For example, a farm manager may be interested in understanding how two irrigation policies compare across multiple features of the future, such as cumulative water usage, plant stress, run off, etc. This can be facilitated by defining reward functions corresponding to each feature and issuing the appropriate PCQs.

\subsection{Learning to Answer PCQs with Confidence}

Given an accurate generative model of the environment MDP, a PCQ $(s,\pi,\hs,\hpi,h)$ can be answered via Monte-Carlo trajectory sampling to estimate $V^{\pi}(s,h)$ and $V^{\hpi}(\hs,h)$. Further, the confidence in the answer can be arbitrarily improved by increasing the number of sampled trajectories. In this work, we do not assume an environment model, but instead are provided with an offline data set of environment trajectories produced by one or more unknown behavior policies. We will denote this dataset by $\mathcal{D}=\{(s_i,a_i,s'_i,r_i)\}$ where each tuple corresponds to an observed transitions from state $s_i$ to state $s'_i$ after taking action $a_i$ and receiving reward $r_i$.

Given a dataset $\mathcal{D}$ we would like to learn a model for predicting answers to PCQs from a query space $\mathcal{Q}$. Here, $\mathcal{Q}$ may assert application-specific restrictions on states and policies involved in PCQs. A fundamental challenge is that the coverage of $\mathcal{D}$ will not necessarily be representative of the dynamics and rewards relevant to answering all queries in $\mathcal{Q}$. Thus, if query answers are being used to inform important decisions, then it is critical for each answer to come with a meaningful measure of confidence that accounts for data coverage and statistical variance. Dealing with this uncertainty is also a core challenge for general offline RL \citep{levine2020offline}, which has lead to a number of approaches for addressing it. However, there is little direct evaluation of the uncertainty-handling components.

The above motivates the \emph{OPCC learning problem}, which provides a dataset $\mathcal{D}$ and desired constraints on the query space $\mathcal{Q}$. The learner should output a model $w=(f,c)$ composed of: 1) a query prediction function $f:\mathcal{Q} \rightarrow \{0,1\}$, which returns a binary answer for any query in $\mathcal{Q}$, and 2) a confidence function $c:\mathcal{Q} \rightarrow [l,u]$ that maps queries in $\mathcal{Q}$ to a confidence value within a bounded interval. Given a query $q$, the intent is for larger values of $c(q)$ to indicate a higher confidence in the prediction $f(q)$. Note that we do not attach any predefined semantics to the values of $c(q)$ to allow for flexibility of potential solutions. Rather, we focus on defining metrics for directly evaluating the quality of uncertainty quantification provided by $w$. If desired, various methods can be used after learning to calibrate the confidence values of $c$ to meaningful scales (e.g. \citep{loh1987,naeini2015}. \cref{sec:baselines} discusses possible learning approach and the baselines evaluated in this paper.

\subsection{Evaluation Metrics}

Since OPCC involves confidence estimation for binary PCQ predictions, we can draw on evaluation metrics from prior work on selective classification (e.g. \citep{el-yaniv2010, xin2021}). In selective classification, the aim is to reduce prediction errors by allowing a predictor to abstain from a prediction if the confidence is below a threshold. \emph{The quality of confidence values is thus related to how well they result in abstaining when the prediction would have been incorrect}. This idea is formalized via \emph{risk-coverage curves (RCCs)}, as outlined below.

\begin{figure}[!h]
\centering
\includegraphics[scale=0.4]{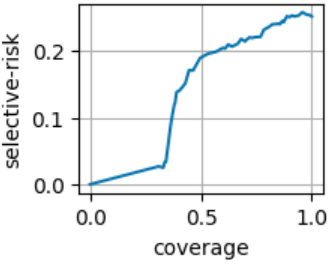}
\caption{A sample of Risk-Coverage Curve (RCC)}
\label{fig:sample-selective-risk-coverage-curve}
\end{figure}

\textbf{Risk-Coverage Curve.} Let $\mathcal{L}(q,\hat{y})$ be a loss function for predicting $\hat{y}$ for query $q$, e.g. 0/1 loss. Given a test set of queries $Q = \{q_1, \ldots, q_N\}$, a model $w=(f,c)$, and confidence threshold $\tau$, the \emph{coverage} is the fraction of test queries with confidence at least $\tau$. The \emph{selective risk} is the average loss of $f$ over the covered queries. Formally, the \emph{coverage} and \emph{selective risk} are respectively define by
\begin{eqnarray}
        cov(w,Q,\tau) &=& \frac{1}{|Q|}\sum_{q\in Q} I[c(q)\geq \tau] \label{eq:coverage} \\
        r(w,Q,\tau) &=& \frac{\sum_{q\in Q} I[c(q)\geq \tau]L(q,f(q))}{\sum_{q\in Q} I[c(q) \geq \tau]} \label{eq:risk}
\end{eqnarray}
\noindent where $I$ is the binary indicator function. Thus, each possible threshold corresponds to a risk-coverage operating point $\langle r(w,Q,\tau), cov(w,Q,\tau)\rangle$. An RCC is simply the risk versus coverage curve of these operating points when sweeping through possible thresholds. Practically, for a finite test set $Q$ there can be at most $|Q|$ unique operating points since there are at most $|Q|$ distinct confidence values produced by $c$. Thus, when displaying empirical RCCs we linearly interpolate between those operating points.\footnote{This is justified by the fact that we can achieve (in expectation) any linearly interpolated operting point between two thresholds $\tau_1$ and $\tau_2$ by varying the probability $p \in [0,1]$ of using $\tau_1$ versus $\tau_2$ to decide on abstention.} Figure \ref{fig:sample-selective-risk-coverage-curve} shows an example of an RCC from our experiments. The curve starts at the point $(0,0)$, since the risk is 0 at zero coverage, and ends at $(1,r_f)$, where $r_f$ is the risk of $f$ evaluated on all of $Q$.

In order to provide a single measure of the RCC quality, we aggregate across all thresholds to compute the \emph{Area Under the RCC }(\textbf{AURCC}). Since lower risk is preferred, we consider a lower AURCC to indicate better confidence estimation. The minimum AURCC is 0, which occurs when the predictor $f$ has zero risk on all of $Q$. Rather, for a ranomized confidence function that returns a uniformly random value in $[l,u]$, the expected AURCC is $r_f$\footnote{This is because for any threshold $\tau$ and uniformly random confidence function, there is a uniform probability of covering any particular query. Thus, the expected risk for any threshold $\tau > l$ is the expected risk over a random draw from the query set, which is $r_f$.}, indicating no ability to quantify prediction uncertainty.

\textbf{RPP.}  Our second selective-classification metric from \citep{xin2021} is \emph{reverse pair proportion (RPP)}. The main idea is that the ordering of confidence values for a pair of queries should reflect the relative prediction loss for those queries. RPP measures how often the confidence value ordering conflicts with the relative losses across all pairs of queries. In particular, a conflict occurs when the loss of $q_1$ is less than $q_2$, but we are more confident about $q_2$ than $q_1$. The RPP is just the fraction of such conflicts. 
\begin{equation*}
        RPP(w,Q) = \frac{1}{|Q|^2}\sum_{q_1,q_2\in Q} I[l(q_1) < l(q_2), c(q_1) < c(q_2)]
\end{equation*}
\noindent where $l(q)=L(q,f(q))$ is the loss of $f$ on $q$.

\textbf{$\boldsymbol{CR_K}$.} Finally, we introduce a new metric on just the confidence function $c$. A practical difference between different confidence functions is the resolution of values that they output in practice. For example, given a set of queries $Q$, one confidence function $c_1$ may result in only three distinct coverage values $cov(w,Q,\tau)$ across all thresholds, while another confidence function $c_2$ results in $|Q|$ distinct coverage values. All else being equal $c_2$ is the preferable function, since it provides a higher level of resolution with respect to abstention/coverage rates. We measure this via \emph{coverage resolution at $K$}, denoted $CR_K$. To compute $CR_K$ for $w=(c,f)$ and query set $Q$, the coverage interval $[0,1]$ is partitioned into $K$ equal bins and we return the fraction of bins which contain $cov(w,Q,\tau)$ for some threshold $\tau$. By increasing $K$ we get a finer grained distinction in coverage resolution.

\section{OPCC Benchmark Construction}
\label{sec:opcc-benchmark-construction}

In this section, we describe our approach to constructing OPCC benchmarks. We first describe our choices of environments, which are based on existing offline RL benchmarks. In particular, the training datasets used in our benchmarks is based on data from those benchmarks. Next we describe our approach to constructing testing query sets for each of the benchmark environments. The outlined benchmark-construction schema is generic, which can be followed by others to extend the set of available OPCC benchmarks. A tabular summary of the benchmark can be found in \cref{sec:opcc-benchmark-summary-appendix}.

\subsection{Environments}

To support easier adoption of our benchmarks, we selected seven environments and corresponding datasets that are currently used in offline RL research. As a first set of OPCC benchmarks, we have chosen to focus on relatively low-dimensional environments with non-image-based observations. This helps focus initial studies on fundamental OPCC capabilities, rather than simultaneously addressing the additional complexities that enter with lower-level perceptual observations such as images.   

\begin{figure}[!h]
    \centering
    \begin{minipage}[b]{0.42\textwidth}
        \includegraphics[width=\textwidth]{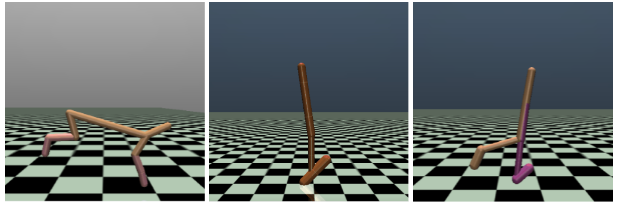}
        \caption{Gym-Mujoco tasks:  half-cheetah, hopper, walker2d (left to right)}
        \label{fig:Gym-Mujoco}
    \end{minipage}
    \hfill
    \begin{minipage}[b]{0.5\textwidth}
        \includegraphics[width=\textwidth]{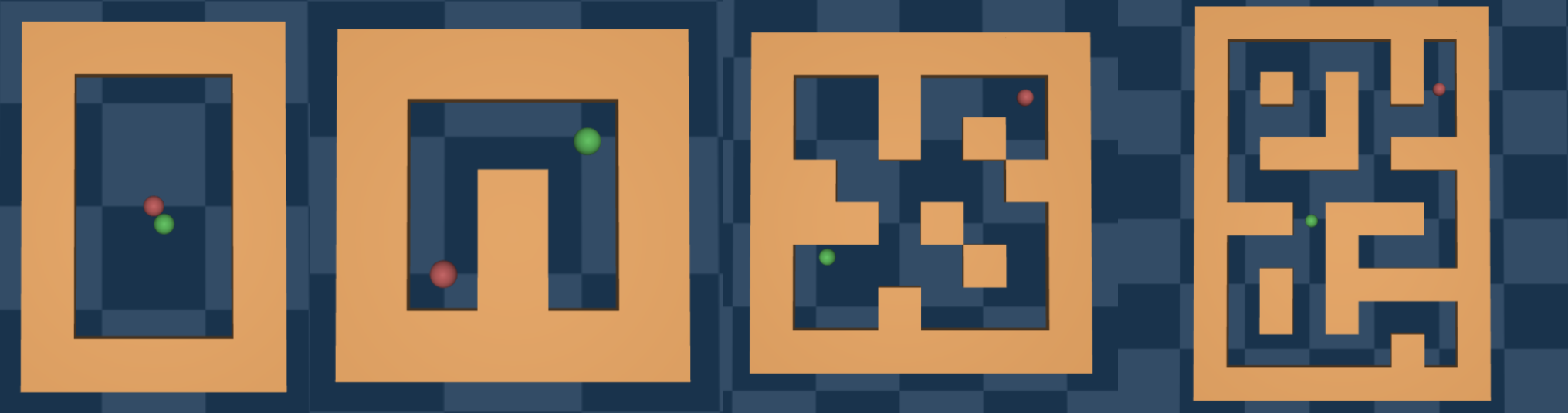}
        \caption{Maze2d tasks: open, umaze, medium, and large ( left to right).}
        \label{fig:Maze2d}
    \end{minipage}
\end{figure}

\textbf{Maze2d (4 environments).} The Maze2d environments were introduced in D4RL \citep{fu2020d4rl} and comprise of 2d mazes of different complexities: \emph{open, u-maze, medium,} and \emph{large} as illustrated in Figure \ref{fig:Maze2d}. Each environment has 4D observations giving the position and velocity of the ball being controlled and a 2D action space specifying the direction of movement. The goal in each environment is to control a rolling ball to reach a goal location. For our benchmarks, we used the sparse-reward version of the environments that provides unit reward for each time step in the goal region. There are no terminal states in these environments and episode ends after maximum number of allowed time-steps reported in \cref{table:opcc-benchmark}. We use the datasets provided by D4RL, which we refer to as \emph{``1M"} due to the datasets each having 1 million state  transitions. The D4RL trajectory data sets were created by running a path-planning algorithm to navigate in the maze between different start and end points.

\textbf{Gym-Mujoco (3 environments).} The Gym-Mujoco environments are based on controlling the actuators of systems within the Mujoco physics simulators. We consider three locomotion-based environments from OpenAI Gym \citep{brockman2016openai}: HalfCheetah, Walker2d, and Hopper. For each environment, we use the corresponding D4RL \citep{fu2020d4rl} datasets that include behavior trajectories of varying qualities. This includes \textit{"random, medium, medium-replay, expert and expert-replay"}. These environments are qualitatively different from Maze2d in that they involve controlling periodic locomotion behavior based on continuous states and actions. Rather Maze2d is primarily about goal-based path planning (navigation) rather than controlling low-level locomotion.

\subsection{Query Set Construction}
For each environment we must create a set of PCQs with ground truth answers for evaluating OPCC. A possible starting point is the off-policy evaluation (OPE) extension \citep{fu2021benchmarks} to D4RL, which includes policies for a subset of the environments. In particular, one of the tasks considered is policy ranking, which is similar in spirit to OPCC. However, that extension of D4RL does not capture at least two important characteristics of OPCC. First, the evaluation protocols do not explicitly address measuring the quality of uncertainty quantification. Of course, this can be addressed by just extending the evaluation protocol and metrics.

Second, and more importantly, it is not clear how to define adequate query sets for evaluating OPCC. In particular, the OPE ranking task from D4RL is currently limited to just ranking policies based on their expected values over the initial state distribution of each environment. In contrast, OPCC evaluations should involve sets of PCQs that cover a wide range of states that are both in-distribution and out-of-distribution relative to the offline data set. Further, it is desirable to select the PCQs in a way that spans some notion of PCQ difficulty. In particular, the notion of difficulty we consider here for a PCQ $(s,\pi,\hs,\hpi,h)$ is directly related to the performance gap between the policies, i.e $\left|V^{\pi}(s,h) - V^{\hpi}(\hs,h)\right|$. It is expected that all else being equal, larger gaps will reason in easier discrimination between policies. Indeed one of the initial challenges in developing the benchmarks was to try to create query sets that were not all too easy or too hard. 

Bases on above considerations we create the evaluation sets of PCQs for each environments via the following steps.  

    {\bf Step 1: Policy Generation.} We first train multiple policies for an environment to serve as the policy used for PCQ construction. We chose to train new policies, rather than use policies from D4RL, to ensure they would be distinct from the behavior policies used to create the D4RL datasets. For each environment we used the corresponding simulator and multiple runs of the PPO algorithm \citep{schulman2017} to train a set of policies of varying quality. We then hand-picked 4 policies with an effort to ensure that they were sufficiently distinct in terms of quality and behavior to support non-trivial PCQs. Performance of these policies in shared in \cref{table:opcc-benchmark-policies} (appendix).

    {\bf Step 2: Initial State Generation.} For each environment we generated a large set of potential initial states by running episodes of the random policy, the learned policies, and a mixture of random and learned policies. This produced a set of states that covered a wide range of the environment that extended well beyond the initial state distributions.

    {\bf Step 3: Candidate PCQ Generation.} For each horizon $h\in \{10,20,30,40,50\}$ we create a  set of 2000 randomly constructed PCQs from the initial states and learned policies. This included explicitly creating random PCQs of the form $(s,\pi,s,\hpi,h)$ with $s$ a random initial state and $\pi,\hpi$ a random pair of the learned policies. In addition, we create a set of PCQs of the form $(s,\pi, \hs, \hpi, h)$ in the same way, except that two random initial states are used instead of one. 

\begin{figure}[!h]
         \centering
         \begin{subfigure}[b]{0.49\textwidth}
             \centering
             \includegraphics[scale=0.54]{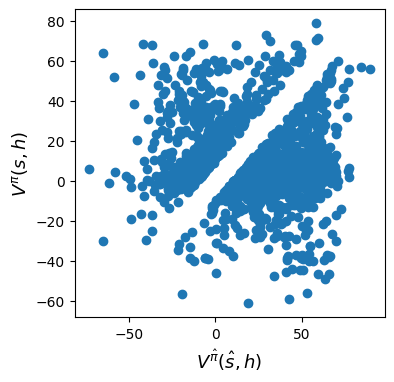}
             \caption{Halfcheetah-v2}
             \label{fig:halfcheetah-v2-query-viz}
         \end{subfigure}
         \hfill
         \begin{subfigure}[b]{0.49\textwidth}
             \centering
             \includegraphics[scale=0.54]{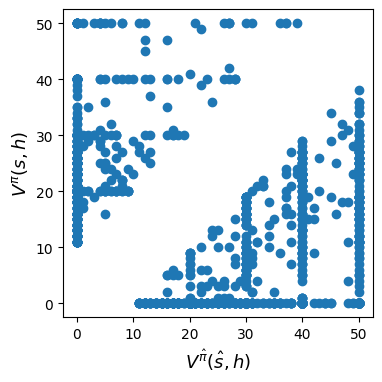}
             \caption{Maze2d-open-v0}
             \label{fig:maze2d-open-v0-query-viz}
         \end{subfigure}
        \caption{Scatter plot of the PCQ in a) HalfCheetah and b) Maze2d-open-v0. For each PCQ $(s,\pi,\hs,\hpi,h)$, we plot $V^{\pi}(s,h)$ vs. $V^{\hpi}(\hs,h)$.}
        \label{fig:query-viz}
    \end{figure}

    {\bf Step 4: PCQ Labeling and Selection.} For each generated PCQ from step 3 we used Monte-Carlo simulation via the environment simulator to accurately estimate $V^{\pi}(s,h)$ and $V^{\hpi}(\hs,h)$ and removed any PCQ having a difference of less than 10  between the value of each side of a query. The motivation is to filter out PCQs that are the most ambiguous and more likely to act as a source of noise in evaluations. Finally, for each $h$, we randomly selected 1500 of the PCQs to include as the benchmark query set. In Figure \ref{fig:query-viz}, we show, a scatter plot of $\left(V^{\pi}(s,h), V^{\hpi}(\hs,h)\right)$ for the selected set of PCQs for Halfcheetah-v2 and maze2d-open-v0. Notice the lack of PCQs along the diagonal, which corresponds to the removal of ambiguous queries. Also note that the PCQs span a range of value gaps, which suggests that they span varying PCQs of varying difficult. If these plots showed a bias toward only large gap queries, then additional steps would be necessary to ensure that more variation was present in the selected query sets. 

\section{OPCC Baselines}
\label{sec:baselines}

In this section, we describe the class of baselines that will be made available with the benchmarks and included in our pilot experiments (\cref{sec:experiments}). Recall that each baseline must provide a prediction function $f$ and confidence function $c$ that are derived from the dataset $\mathcal{D}$. Perhaps the most natural approach for $f$ to answer a PCQ $(s,\pi,\hs,\hpi,h)$ is to estimate and then compare the relevant values using OPE. The corresponding confidence function $c$ might then be based on the uncertainty of the value estimates.

There are at least two types of OPE approaches to consider: model-free and model-based. Model-free approaches, such as fitted Q-evaluation \citep{ernst2005tree} typically learn a Q-function $Q^{\pi}(s,a)$ for a given policy $\pi$ that can be evaluated for any state and action.
Unfortunately, each function learned by such model-free methods is valid for the single policy $\pi$ and the effective horizon used during training. Thus, answering PCQs involving other policies or horizons requires costly retraining. Since we are seeking an OPCC approach, which can be quickly applied to arbitrary policies, states, and horizons, we instead choose to use a model-based approach for our baselines.

Our baselines are variants of model-based ensemble approaches, which are one of the most common class of approaches used in model-based RL for dynamics modeling and capturing uncertainty \citep{argenson2020model,yu2020mopo,kidambi2020morel}. In particular, our baselines all have the following structure: 
\begin{enumerate}
    \item Learn an ensemble of models $\{\hP_i\}$ from $\mathcal{D}$ that each predict the dynamics and reward of the environment.
    \item Use each model in the ensemble to generate estimates of the relevant PCQ values, $V^{\pi}(s,h)$ and $V^{\hpi}(\hs,h)$, via Monte-Carlo simulation of the policies.
    \item Combine the ensemble estimates to provide a prediction and confidence value. 
\end{enumerate}
Compared to model-free approaches, this approach can instantly apply to arbitrary policies and horizons without costly retraining. That is, new policies and horizons can easily be swapped into the Monte-Carlo simulation of step 2 with no modifications to the model. We can obtain different baselines by varying the choices for learning the model ensemble (step 1) as well as varying the ensemble combination approach (step 3). Below we describe the variations used in our experiments. 


\subsection{Base Models} 

In our experiments, we consider two types of base models for forming ensembles. 
The first base model is the commonly use \emph{Feed-Forward (FF)} Gaussian model, which, given the current state/observation and action as input, returns the mean and diagonal covariance matrix of a Gaussian distribution over the next state and reward.
This model allows for stochastic Monte-Carlo simulations by drawing the next state from the model's Gaussian distribution at each time step. In this work, we use the same FF base-model architecture and training details as MBPO \citep{janner2019trust}.



We also consider a recent base model \citep{zhang2021autoregressive} (referred to as \emph{Auto-regressive (AR)}), which was demonstrated in some cases to improve over the output architecture of FF. Instead of generating all $n$ features of the predicted next state in a single pass, AG auto-regressively samples each feature one at a time using $n$ forward passes. In particular, to sample state feature $i$ of the next state, denoted $s_{t+1}^i$, the network receives the usual input $s_t$ and $a_t$ as well as the previously sampled state features $s^{0}_{t+1},... s^{i-1}_{t+1}$. AR then returns the mean and covariance for a Gaussian that is used to sample $s_{t+1}^i$. The intuition is that this approach may allow for representing non-Gaussian and multi-modal next-state distributions compared to the uni-modal Gaussian FF model.


\subsection{Ensemble Learning} 

Model-based approaches to ORL have commonly used ensembles as an attempt to quantify uncertainty, e.g. via measures of ensemble-member disagreement \citep{janner2019trust}. We consider two choices for generating ensembles. The first choice is the standard \emph{bootstrapping ensemble} approach, which simply trains each ensemble member using a different random weight initialization and bootstrapped dataset $\hat{\mathcal{D}}$ by sampling from $\mathcal{D}$ with replacement $|\mathcal{D}|$ times. The intent is that the combination of classic statistical bootstrapping \cite{efron1994} and random initialization will produce a diverse set of ensemble models. 

Often, however, it is observed that the basic bootstrapping approach does not create enough diversity in an ensemble, which is counter to our motivation of representing uncertainty. For this reason, there are a number of proposals for increasing the ensemble diversity, of which, we consider just one in this work. In particular, work motivated by capturing uncertainty in ORL proposed the use of \emph{randomized constant priors} to increase ensemble diversity \citep{osband2018randomized}. For each base model, a randomized constant prior is produced, which is simply a network with random initial weights. The base model is trained as an additive component on top of this prior and the final output is the sum of the two. The intuition is that the constant prior should cause ensemble members to disagree more often in unrepresented parts of the state-space, which will provide a better measure of disagreement-based uncertainty.

\subsection{Prediction and Confidence Values} 

Given a PCQ $(s,\pi,\hs,\hpi,h)$ query and ensemble of size $M$ we generate a prediction and confidence by first using each ensemble member to generate, via Monte-Carlo simulation, a pair of value estimates of $V^{\pi}(s,h)$ and $V^{\hpi}(\hs,h)$. This results in a set of $M$ value estimate pairs, denoted by $\mathcal{V} = \{(V_1,\hV_1), \ldots, (V_M,\hV_M)\}$. This approach is based on the classic view, from bagging classifiers \citep{breiman1996}, of the base learning algorithm being a stochastic function\footnote{Indeed, each run of the base algorithm has at least two sources of randomness in our implementation. First, each run uses a different bootstrap sample of the dataset. Second, each run uses randomized initial weights. Third, mini-batches in stochastic gradient descent depend on the random seed}. The set $\mathcal{V}$ can then be viewed as value-estimate pairs sampled from the distribution of learning algorithm runs. Bagging analysis tells us that if 50\% of the learning algorithm runs result in estimates $(V_i,\hV_i)$ that correctly rank the values, then a large enough ensemble will correctly predict the query.\footnote{This argument assumes independence of the ensemble members, which clearly is not true in practice due to at least correlation between their training data.} Based on this view, below we describe the three approaches we consider for producing predictions and confidences from $\mathcal{V}$. 


\begin{itemize}
\item \emph{Ensemble Voting (EV).} Following \citep{dietterich2000ensemble}, EV simply returns a prediction for a query based on the majority vote across the ensemble of $V_i < \hV_i$. The confidence score is equal to the fraction of ensemble members that agree on the majority vote (in the range [0.5,1]), but re-scaled to fall in the range [0,1].




\item \emph{Paired Confidence Interval (PCI).} The PCI confidence value is computed by estimating the expected value of $V-\hV$ for a random run of the learning algorithm. The mean estimate is given by $\sum_i V_i - \hV_i$ and the prediction is based on the sign of this estimate. The confidence value is based on computing $\alpha$ percentile confidence intervals on the difference, denoted by $[l_{\alpha}, u_{\alpha}]$. In particular, it is equal to the largest value of $\alpha$ such that $0 \not\in [l_{\alpha}, u_{\alpha}]$. Thus, a high confidence value reflects that there is strong evidence that the expected difference is either above or below zero (in agreement with the prediction). Confidence intervals are computed based on the $t$ distribution.



\item \emph{UnPaired Confidence Interval (U-PCI).} This approach makes the prediction in the same way as PCI, but uses unpaired confidence intervals to compute the confidence, which should be expected to be more conservative. In particular, we compute $\alpha$ percentile confidence intervals for the means of the $V_i$ and $\hV_i$ denoted respectively by $\left[l_{\alpha},u_{\alpha}\right]$ and $\left[\hat{l}_{\alpha},\hat{u}_{\alpha}\right]$ and let the confidence be the maximum value of $\alpha$ for which the confidence intervals do not overlap.



\end{itemize}

\section{Related Work}

\textbf{Dynamics Learning in RL.} There has been much recent interest in learning deep models of dynamical systems to support model-based RL. Examples from online RL include \citet{clavera2018model, kurutach2018model}, which learn one-step observation-based dynamics along with extensions to ensembles \citet{deisenroth2011pilco, chua2018deep, janner2019trust,nagabandi2020deep}. PILCO \citep{deisenroth2011pilco,gal2016improving} is another model-based RL approach that learns dynamics via Gaussian Processes  \citep{rasmussen2003gaussian}, which are able to capture epistemic uncertainty. However, performance is primarily measured in terms of overall task performance and it is unclear how well uncertainty is actually quantified. Recent work on offline reinforcement, such as MBOP\citep{argenson2020model}, MOPO \citep{yu2020mopo}, and MoREL \citep{kidambi2020morel} has also considered learning dynamics models over observations from fixed, offline data sets. These approaches incorporate uncertainty estimates in different ways (e.g. pessimistic rewards or dynamics) and all use ensembles to estimate uncertainty. Thus, they are established exemplars of the baselines considered in our work. However, only the final task performance is tested and it is unclear how well uncertainty is actually captured by the models. COMBO \citep{yu2021combo}, Muzero Unplugged \citep{schrittwieser2021online}, and LOMPO \citep{rafailov2021offline} investigate learning latent-space dynamics models for offline RL rather than learning in the observation space. Again, however, there is no explicit evaluation of the models ability to quantify uncertainty.

\textbf{Policy Ranking.} Similar to our work, \citet{sonabend2020expert} also identifies the need to measure uncertainty for policies learned with limited data.
In order to learn safe policies, their approach uses hypothesis testing for  determining uncertainty in policy evaluation for a pair of candidate policies based on sampling from model posteriors. This helps in ranking them and selection of better performing policy over the behavior policy in a safe manner. The work, however, was limited to small flat state-spaces and did not explicitly evaluate uncertainty quantification. 
In contrast, our work produces a benchmark to primarily focus on uncertainty quantification of a system using offline data, rather than evaluating in terms of overall task performance.  

DOPE \citep{fu2021benchmarks} studies OPE and devises a protocol that measures policy evaluation, ranking, and selection. For this purpose, the approach introduces a set of candidate policies along with their expected value over a distribution of initial states. Rather, in our work, we question the ability of a system to rank policies from any arbitrary state for a given horizon instead of limiting to initial state distribution only. This can help provide a more comprehensive view of uncertainty estimation across the state space.

In similar motivation, SOPR-T\citep{jin2021supervised} also considers policy ranking from offline data and additional policy-value supervision.
This is done by learning an encoded representation of a policy using a transformer based architecture and a scoring function over the representation. In order to learn the representation, they require a set of pre-defined policies, each labeled by its ground truth value with respect to an initial state distribution. Our framework does not assume the availability of such policy-value supervision and also puts an emphasis on uncertainty quantification, which is not evaluated by this work.


\textbf{Confidence Intervals.} We make use of confidence intervals over policy value estimates for answering queries. \citet{thomas2015high} also studies confidence interval estimation over policy value estimates using trajectories generated by a different set of policies. Their approach uses importance sampling for unbiased value estimates, which suffers from high variance leading to loose confidence bounds. They also introduce the problem of \emph{high confidence off-policy evaluation} and produce tighter bounds on estimates using improved concentration inequalities \citep{massart2007concentration}. 

Another class of approaches is based on statistical bootstrapping \citep{efron1987better}. \cite{hanna2017bootstrapping} bootstraps learned MDP transition models in order to estimate lower confidence bounds on policy evaluation estimates with limited data. \citet{kostrikov2020statistical} suggests that confidence intervals of these bootstrapped estimates are not guaranteed to be accurate. In practice, they are shown to be overly confident  especially for insufficient sample sizes and under-coverage of the data distribution. They suggest, that, in practice, this issue may be mitigated by inducing noisy rewards and regularization to learn smoother empirical transition and reward functions. Evaluating that claim within our OPCC framework is a potential direction for future work.  

CoinDICE \citep{dai2020coindice}, and similar methods \citep{nachum2019dualdice, zhang2020gradientdice, strehl2008analysis,hao2021bootstrapping} progressively focus on confidence intervals for OPE based on the formulation of certain optimization problems. 
This iterative optimization  approach for estimating policy value and confidence bounds induces a computational overload. This is undesirable in our framework which aims to rapidly answer queries of arbitrary horizons and policies making it computationally unsustainable. An interesting direction for future work is to consider generalizing this optimization-based approach to more flexibly handle arbitrary policies.

\section{Experiments}
\label{sec:experiments}

Our pilot experiments explore the baseline methods on our benchmarks using the proposed metrics for OPCC. It is important to note that these experiments are not intended to identify a top performer. Rather our primary goal for these pilot experiments is to assess the adequacy of the benchmarks and metrics for future work and to establish a basic performance bar. Secondarily, we are interested to observe evidence or the lack of evidence for certain assumptions that might be drawn about the baselines from prior work. Based on those primary goals our experiments and analysis are designed to: 1) Assess whether the benchmarks appear to be too difficult or too easy for supporting future work; 2) Assess whether there is any evidence that our baselines are sensitive to the data-set type used for each benchmark environment. In particular, the performance of strong OPCC approach should vary with the coverage afforded by the data set. 3) Assess whether there is any evidence for performance differences among the baseline variations.
In particular, certain features such as auto-regressive sampling, constant priors, and larger ensembles have been claimed to improve uncertainty handling in prior work. 

In our experiments, unless otherwise specified the default model is an ensemble of 100 deterministic feed-forward models and uses EV for the confidence score. 
Two additional details are important to note for these experiments. First, as is customary in model-based RL (including ORL), we are using a pre-defined episode termination function rather than a learned one. We have found that this can significantly impact performance of model-based RL systems and also our OPCC evaluations. Second, we clipped predicted observations and rewards to keep them within the bounds of the available data sets, which is also a common practice in ORL that we found to be important.

\begin{figure}[h!]
\centering
\includegraphics[scale=0.6]{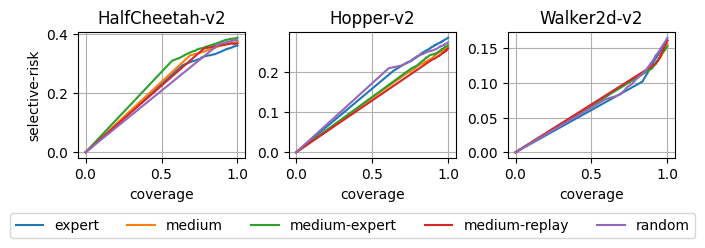}
\caption{Selective-risk coverage curves for different \emph{gym-mujoco} environements and \emph{dataset types} (depicted by different colors). The x-axis spans from no(0) coverage to complete(1) coverage of queries and the y-axis is the risk for the corresponding query coverage. Each risk-coverage point is determined by varying the confidence threshold.} 
\label{fig:gym-mujoco-dataset-quality}
\end{figure}
\begin{figure}[!ht]
\centering
\includegraphics[scale=0.6]{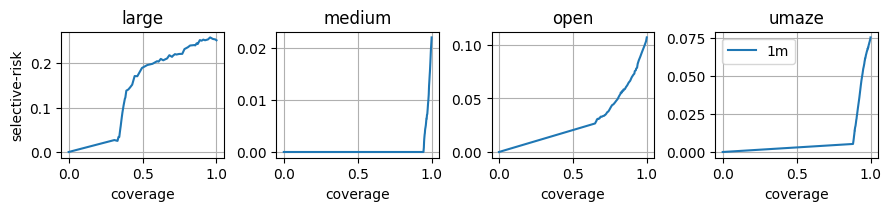}
\caption{Selective-risk coverage curve for `1m' dataset in \emph{maze} environments. This is the complete navigation dataset of \emph{1 million} transactions.} 
\label{fig:maze-dataset-quality}
\end{figure}

{\bf Too hard or too easy?} We first assess the degree of difficulty posed by our OPCC benchmark for our baselines. \cref{fig:gym-mujoco-dataset-quality,fig:maze-dataset-quality} show RCCs of our default model for  different data set types (averaged across the different PCQ horizons $h$) in gym-mujoco and maze2d environments. \cref{table:gym-mujoco-dataset-quality,table:maze-dataset-quality} report their corresponding metrics i.e. AURCC, RPP, $CR_k$, and Loss (or risk) at complete coverage.

First, we consider risk at complete coverage and find that there is no significant difference in risk across dataset type, but varies significantly across gym-mujoco environments. This shows that some environments are more challenging than others due to their underlying complex dynamics and high dimensional observation and action sizes. Also, the risk at complete coverage for maze2d environments with a single dataset ('1m') is significantly lower than gym-mujoco. This is potentially due to data collection via a path-planning procedure leading to significant state-action space coverage. Further Medium and Umaze have very small risks without much room for risk improvement, while Large and Open appear to have room for improvement. Second, we consider how the risk varies across coverage values. In most cases, there are no thresholds that produce points within the coverage interval (0,0.5], which indicates a lack of sensitivity in that coverage range. There are typically multiple points between (0.5, and 1], though often just a few. Ideally we would hope for a more gradual degradation in risk spanning from no coverage to complete coverage. This suggests that there is  significant room to improve the coverage sensitivity, especially in the range [0,0.5].

Overall, the current set of benchmarks, with the exception of 2 Maze2d environments, are not too easy and appears to offer significant room for improvement in terms of both overall risk and sensitivity of the RCCs across coverage values. Likewise, the observation that the risks achieved are significantly less than chance suggest that the benchmarks are not too hard. 

{\bf Impact of dataset type.} The different types of data sets provide different types of coverage of the system dynamics. Is there evidence that our baselines are able to distinguish among these types? \cref{fig:gym-mujoco-dataset-quality} shows that the RCCs for different datasets are quite similar for each of the gym-mujoco environment. The AURCC and RPP values in  \cref{table:gym-mujoco-dataset-quality} are consistent with these observations. This could be due to the diverse coverage of queries across the state space that offer challenges for all datasets. The small variation in RCCs across dataset types could also be due to the models learned from different datasets providing similar types of generalization. It is also possible that differences between dataset types would become more prevalent for smaller versions of the datasets, which an interesting future extension to the benchmarks. Finally no significant patterns for CR in relation to data-set type are apparent, which is not surprising since CR is expected to be more heavily influenced by the type of baseline approach.

\begin{table}[!ht]
\caption{Evaluation metrics for  \emph{dataset-type} comparison in \emph{gym-mujoco} environments. This includes mean and confidence intervals estimates at $95\%$ confidence level for metrics corresponding to 5(seed) dynamics trained over each dataset.} 
\label{table:gym-mujoco-dataset-quality}
\begin{center} 
\begin{footnotesize} 
\begin{sc}
    \begin{tabular}{|l | c | c | c| c | c | c |}
        \toprule
        Env. & dataset-type & AURCC$(\downarrow)$ & RPP$(\downarrow)$ & $CR_K(\uparrow)$ &  loss$(\downarrow)$ \\
        \midrule
        \multirow{5}{3.6em}{Half\\Cheetah}  & expert & $0.212\pm0.002$ & $0.05\pm0.001$ & $0.5\pm(<0.001)$ & $0.361\pm0.002$ \\
        & medium & $0.222\pm0.001$ & $0.048\pm0.001$ & $0.5\pm(<0.001)$ & $0.374\pm0.002$ \\
        & medium-expert & $0.24\pm0.004$ & $0.06\pm0.002$ & $0.6\pm(<0.001)$ & $0.387\pm0.003$ \\
        & medium-replay & $0.216\pm0.001$ & $0.04\pm0.001$ & $0.4\pm(<0.001)$ & $0.368\pm0.001$ \\
        & random & $0.206\pm0.001$ & $0.023\pm0.001$ & $0.3\pm(<0.001)$ & $0.378\pm0.001$ \\
        \midrule
        \multirow{5}{3.6em}{Hopper}  & expert & $0.152\pm0.002$ & $0.04\pm0.001$ & $0.5\pm(<0.001)$ & $0.284\pm0.002$ \\
        & medium & $0.133\pm0.001$ & $0.03\pm0.001$ & $0.4\pm(<0.001)$ & $0.26\pm0.002$ \\
        & medium-expert & $0.136\pm0.001$ & $0.028\pm(<0.001)$ & $0.4\pm(<0.001)$ & $0.265\pm0.001$ \\
        & medium-replay & $0.128\pm0.001$ & $0.012\pm0.001$ & $0.3\pm(<0.001)$ & $0.258\pm0.001$ \\
        & random & $0.156\pm0.008$ & $0.045\pm0.004$ & $0.54\pm0.043$ & $0.273\pm(<0.001)$ \\
        \midrule
        \multirow{5}{3.6em}{Walker\\2d}  & expert & $0.064\pm0.001$ & $0.011\pm(<0.001)$ & $0.3\pm(<0.001)$ & $0.161\pm(<0.001)$ \\
        & medium & $0.069\pm0.001$ & $0.007\pm(<0.001)$ & $0.22\pm0.035$ & $0.156\pm0.001$ \\
        & medium-expert & $0.068\pm(<0.001)$ & $0.007\pm(<0.001)$ & $0.24\pm0.043$ & $0.153\pm0.001$ \\
        & medium-replay & $0.07\pm0.001$ & $0.005\pm(<0.001)$ & $0.2\pm(<0.001)$ & $0.161\pm0.001$ \\
        & random & $0.067\pm0.001$ & $0.024\pm0.001$ & $0.54\pm0.043$ & $0.165\pm0.001$ \\
        \bottomrule
    \end{tabular}
\end{sc}
\end{footnotesize} 
\end{center} 
\end{table}

\begin{table}[!ht]
\caption{Evaluation metrics for  \emph{dataset-types} comparison in \emph{maze} environments. This includes mean and confidence intervals estimates at $95\%$ confidence level for metrics corresponding to 5(seed) dynamics trained over each dataset.} 
\label{table:maze-dataset-quality}
\begin{center} 
\begin{footnotesize} 
\begin{sc}
    \begin{tabular}{|l | c | c | c| c | c | c |}
        \toprule
        Env. & dataset-type & AURCC$(\downarrow)$ & RPP$(\downarrow)$ & $CR_K(\uparrow)$ &  loss$(\downarrow)$ \\
        \midrule
        large & 1m & $0.14\pm0.015$ & $0.062\pm0.004$ & $0.82\pm0.035$ & $0.251\pm0.029$ \\
        \midrule
        medium & 1m & $0.001\pm(<0.001)$ & $0.0\pm(<0.001)$ & $0.2\pm(<0.001)$ & $0.022\pm0.001$ \\
        \midrule
        open & 1m & $0.029\pm0.001$ & $0.012\pm0.001$ & $0.5\pm(<0.001)$ & $0.107\pm0.005$ \\
        \midrule
        umaze & 1m & $0.008\pm0.001$ & $0.002\pm(<0.001)$ & $0.3\pm(<0.001)$ & $0.075\pm0.003$ \\
        \bottomrule
    \end{tabular}
\end{sc}
\end{footnotesize} 
\end{center} 
\end{table}


{\bf Impact of Query Horizon.} Learned dynamics are well known to suffer from error accumulation in multi-step rollouts. This leads to the hypothesis that OPCC performance might degrade with increasing query horizons. In \cref{table:gym-mujoco-horizon} and \cref{table:maze-horizon} (Appendix) we provides metrics for various horizons $h$ averaged across data-set types. As expected, we observe higher AURCCs for longer horizons, which provides positive evidence for the hypothesis.  Interestingly, we observe that in most of the environments we have very low risk for short horizons. In general, we  observe AURCCs for $h=10$ or $h=20$ are at least an order of magnitude smaller than for larger horizons across the benchmark. This suggests a possible threshold effect for OPCC with respect to increasing horizon due to error accumulation. It also suggests our current baselines are better suited for applications like reliable policy improvement with smaller horizons.

\begin{table}[tb!]
\caption{Evaluation metrics for  \emph{horizon} comparison in \emph{gym-mujoco} environments. These mean and confidnce interval(95$\%$) estimates are over 50 samples corresponding to 5(seed) dynamics trained over 5 different datasets.} 
\label{table:gym-mujoco-horizon}
\begin{center} 
\begin{footnotesize} 
\begin{sc}
    \begin{tabular}{|l | c | c | c| c | c | c |}
        \toprule
        Env. & horizon & AURCC$(\downarrow)$ & RPP$(\downarrow)$ & $CR_K(\uparrow)$ &  loss$(\downarrow)$ \\
        \midrule
        \multirow{5}{3.6em}{Half\\Cheetah}  & 10.0 & $0.077\pm0.004$ & $0.008\pm0.001$ & $0.288\pm0.017$ & $0.191\pm0.006$ \\
        & 20.0 & $0.217\pm0.006$ & $0.041\pm0.005$ & $0.416\pm0.042$ & $0.374\pm0.005$ \\
        & 30.0 & $0.215\pm0.004$ & $0.038\pm0.005$ & $0.404\pm0.038$ & $0.377\pm0.005$ \\
        & 40.0 & $0.223\pm0.006$ & $0.049\pm0.006$ & $0.464\pm0.041$ & $0.368\pm0.005$ \\
        & 50.0 & $0.277\pm0.008$ & $0.063\pm0.007$ & $0.516\pm0.054$ & $0.428\pm0.003$ \\
        \midrule
        \multirow{5}{3.6em}{Hopper}  & 10.0 & $0.017\pm0.002$ & $0.001\pm(<0.001)$ & $0.2\pm(<0.001)$ & $0.048\pm0.002$ \\
        & 20.0 & $0.078\pm0.002$ & $0.016\pm0.003$ & $0.336\pm0.038$ & $0.17\pm0.003$ \\
        & 30.0 & $0.146\pm0.004$ & $0.029\pm0.004$ & $0.42\pm0.033$ & $0.284\pm0.005$ \\
        & 40.0 & $0.169\pm0.007$ & $0.038\pm0.006$ & $0.432\pm0.036$ & $0.293\pm0.004$ \\
        & 50.0 & $0.196\pm0.009$ & $0.047\pm0.007$ & $0.516\pm0.049$ & $0.334\pm0.006$ \\
        \midrule
        \multirow{5}{3.6em}{Walker\\2d}  & 10.0 & $0.011\pm(<0.001)$ & $0.001\pm(<0.001)$ & $0.22\pm0.016$ & $0.033\pm0.003$ \\
        & 20.0 & $0.025\pm0.002$ & $0.003\pm0.001$ & $0.252\pm0.042$ & $0.077\pm0.004$ \\
        & 30.0 & $0.059\pm0.001$ & $0.01\pm0.003$ & $0.3\pm0.061$ & $0.132\pm0.004$ \\
        & 40.0 & $0.093\pm0.002$ & $0.017\pm0.004$ & $0.384\pm0.049$ & $0.209\pm0.004$ \\
        & 50.0 & $0.131\pm0.002$ & $0.023\pm0.005$ & $0.392\pm0.06$ & $0.259\pm0.002$ \\
        \bottomrule
    \end{tabular}
\end{sc}
\end{footnotesize} 
\end{center} 
\end{table}




\vspace{-0.4cm}
{\bf Influence of different confidence functions.} \cref{table:gym-mujoco-uncertainty-type} and \cref{table:maze-uncertainty-type}(Appendix) gives metrics for our three different uncertainty functions (EV, PCI, and U-PCI) averaged over data-set types and horizons. The results for AURCC and RPP both indicate evidence that the confidence interval approaches (PCI and U-PCI) have an advantage over EV. This is encouraging as it suggests considering other more sophisticated statistical testing approaches may lead to further improvement. However, the results for CR indicate that the confidence interval approaches have significantly less resolution than EV. This may lead to poorer performance for probability calibration approaches applied to PCI or U-PCI confidence scores. Further work is required to understand this decrease in resolution.



\begin{table}[tb!]
\caption{Evaluation metrics for  \emph{uncertainty-type} comparison in \emph{gym-mujoco} environments.  These mean and confidence interval(95$\%$) estimates are over 50 samples corresponding to 5(seed) dynamics trained over 5 different datasets.} 
\label{table:gym-mujoco-uncertainty-type}
\begin{center} 
\begin{footnotesize} 
\begin{sc}
    \begin{tabular}{|l | c | c | c| c | c | c |}
        \toprule
        Env. & uncertainty-type & AURCC$(\downarrow)$ & RPP$(\downarrow)$ & $CR_K(\uparrow)$ &  loss$(\downarrow)$ \\
        \midrule
        \multirow{3}{3.6em}{Half\\Cheetah}  & ev & $0.219\pm0.005$ & $0.044\pm0.005$ & $0.46\pm0.04$ & $0.374\pm0.004$ \\
        & pci & $0.191\pm0.002$ & $0.006\pm0.001$ & $0.2\pm(<0.001)$ & $0.373\pm0.004$ \\
        & u-pci & $0.196\pm0.003$ & $0.014\pm0.002$ & $0.228\pm0.018$ & $0.373\pm0.004$ \\
        \midrule
        \multirow{3}{3.6em}{Hopper}  & ev & $0.141\pm0.005$ & $0.031\pm0.005$ & $0.428\pm0.034$ & $0.268\pm0.004$ \\
        & pci & $0.135\pm0.002$ & $0.004\pm0.001$ & $0.2\pm(<0.001)$ & $0.269\pm0.004$ \\
        & u-pci & $0.135\pm0.003$ & $0.009\pm0.002$ & $0.216\pm0.014$ & $0.269\pm0.004$ \\
        \midrule
        \multirow{3}{3.6em}{Walker\\2d}  & ev & $0.068\pm0.001$ & $0.011\pm0.003$ & $0.3\pm0.051$ & $0.159\pm0.002$ \\
        & pci & $0.078\pm0.001$ & $0.002\pm0.001$ & $0.2\pm(<0.001)$ & $0.16\pm0.003$ \\
        & u-pci & $0.076\pm0.001$ & $0.004\pm0.001$ & $0.22\pm0.016$ & $0.16\pm0.003$ \\
        \bottomrule
    \end{tabular}
\end{sc}
\end{footnotesize} 
\end{center} 
\end{table}


{\bf Impact of Ensemble Size.} We now consider the impact of ensemble size for our baselines. \cref{table:gym-mujoco-ensemble-count} and \cref{table:maze-ensemble-count} (Appendix) show the results for ensemble sizes ranging from 10 to 100. Our prior expectation was that performance would increase with significant increase in ensemble-size. In general, we do not see statistically significant differences between ensemble sized for AURCC based on our current experimental budget (i.e. confidence intervals intersect).  However, based on trends in the means, there is weak evidence of improved AURCC. The exception is HalfCheetah, where for AURCC, the trends is opposite of the expectation. However, the differences in means tends to be small, suggesting that ensemble size is not having a large impact even if more computational budget were devoted to support statistical significance. 

For RPP and Coverage Resolution ($CR_k$) there is typically a statically significant improvement from ensemble size 10 to 100. The exceptions are umaze and medium-maze where losses are very small for all ensemble sizes. Overall, however, differences are relatively small in magnitude. This may be due to the ensembles not being diverse enough, or the base models used to construct the ensembles are not accurate enough. These results demonstrate the value of the OPCC benchmarks in being able to explicitly test hypotheses about uncertainty quantification, rather than relying on downstream results that may be impacted by many possible factors. 

\textbf{Randomized Constant Priors}. In order to encourage diversity, we introduce \emph{randomized constant priors} in our ensemble models. These are suggested to encourage extrapolation diversity, especially on out-of-distribution state-action pairs, which could improve disagreement-based uncertainty estimates.  However, when we included the constant priors in our model, we didn't find significant improvements in our evaluation metrics as shown in \cref{table:maze-prior-scale}(Appendix) and \cref{table:gym-mujoco-prior-scale}.
We use the same architecture as the dynamics model for \emph{prior} with random weights and scale them with \emph{``prior-scale"} before adding them to ensemble models and a prior-scale of 0 indicates no usage of them. In the case of maze2d, we generally observe a slight (but statistically insignificant) reduction in \emph{AURCC}, whereas \emph{RPP} and $CR_K$ tends to remains same. On the contrary, in the case of gym-mujoco (\cref{table:gym-mujoco-prior-scale}), we generally observe a slight (but statistically insignificant) increase in \emph{AURCC} , \emph{RPP} and $CR_K$.

Prior work by \citet{osband2018randomized} demonstrated improvement in end-task RL performance by having an ensemble of DQN \citep{mnih2013playing} models with randomized constant priors. However, explicit analysis of the uncertainty quantification was not provided. Our observations suggests that randomized constant priors do not appear to improve uncertainty quantification at least as measured through our OPCC benchmarks. Further investigation is necessary to better understand the performance differences observed in \citet{osband2018randomized}. An interesting direction of future work is to consider other previously proposed mechanisms for improving ensemble diversity within the OPCC framework. 

\begin{table}[t!]
\caption{Evaluation metrics for  \emph{ensemble-count} comparison in \emph{gym-mujoco} environments. We train 5(seed) ensemble dynamics of size 100 for each dataset and start with ensemble of 10 models for OPCC metrics estimation. Thereafter, we incremently increase their exposure for  metrics mean and confidence intervals (95$\%$) } 
\label{table:gym-mujoco-ensemble-count}
\begin{center} 
\begin{footnotesize} 
\begin{sc}
    \begin{tabular}{|l | c | c | c| c | c | c |}
        \toprule
        Env. & ensemble-count & AURCC$(\downarrow)$ & RPP$(\downarrow)$ & $CR_K(\uparrow)$ &  loss$(\downarrow)$ \\
        \midrule
        \multirow{5}{3.6em}{Half\\Cheetah}  & 10 & $0.209\pm0.004$ & $0.028\pm0.004$ & $0.32\pm0.029$ & $0.378\pm0.004$ \\
        & 20 & $0.213\pm0.004$ & $0.034\pm0.004$ & $0.36\pm0.04$ & $0.376\pm0.004$ \\
        & 40 & $0.215\pm0.004$ & $0.039\pm0.005$ & $0.4\pm0.035$ & $0.374\pm0.004$ \\
        & 80 & $0.219\pm0.005$ & $0.043\pm0.005$ & $0.452\pm0.04$ & $0.374\pm0.004$ \\
        & 100 & $0.219\pm0.005$ & $0.044\pm0.005$ & $0.46\pm0.04$ & $0.374\pm0.004$ \\
        \midrule
        \multirow{5}{3.6em}{Hopper}  & 10 & $0.141\pm0.005$ & $0.02\pm0.004$ & $0.316\pm0.029$ & $0.273\pm0.005$ \\
        & 20 & $0.141\pm0.005$ & $0.024\pm0.004$ & $0.344\pm0.037$ & $0.272\pm0.004$ \\
        & 40 & $0.141\pm0.005$ & $0.027\pm0.004$ & $0.388\pm0.04$ & $0.27\pm0.004$ \\
        & 80 & $0.141\pm0.005$ & $0.03\pm0.004$ & $0.416\pm0.038$ & $0.269\pm0.004$ \\
        & 100 & $0.141\pm0.005$ & $0.031\pm0.005$ & $0.428\pm0.034$ & $0.268\pm0.004$ \\
        \midrule
        \multirow{5}{3.6em}{Walker\\2d}  & 10 & $0.072\pm0.001$ & $0.007\pm0.002$ & $0.256\pm0.03$ & $0.16\pm0.002$ \\
        & 20 & $0.071\pm0.001$ & $0.008\pm0.002$ & $0.268\pm0.038$ & $0.16\pm0.002$ \\
        & 40 & $0.07\pm0.001$ & $0.01\pm0.003$ & $0.28\pm0.046$ & $0.16\pm0.002$ \\
        & 80 & $0.068\pm0.001$ & $0.01\pm0.003$ & $0.284\pm0.049$ & $0.159\pm0.002$ \\
        & 100 & $0.068\pm0.001$ & $0.011\pm0.003$ & $0.3\pm0.051$ & $0.159\pm0.002$ \\
        \bottomrule
    \end{tabular}
\end{sc}
\end{footnotesize} 
\end{center} 
\end{table}

\begin{table}[tb!]
\caption{Evaluation metrics for  \emph{prior-scale} comparison in \emph{gym-mujoco} environments comprising of mean and confidence interval(95$\%$) over 50 samples belonging to 5(seed) dynamics models for each of the 5 datasets. Prior scale of 0 means no randomized constant prior is added. } 
\label{table:gym-mujoco-prior-scale}
\begin{center} 
\begin{footnotesize} 
\begin{sc}
    \begin{tabular}{|l | c | c | c| c | c | c |}
        \toprule
        Env. & prior-scale & AURCC$(\downarrow)$ & RPP$(\downarrow)$ & $CR_K(\uparrow)$ &  loss$(\downarrow)$ \\
        \midrule
        \multirow{2}{3.6em}{Half\\Cheetah}  & 0 & $0.219\pm0.005$ & $0.044\pm0.005$ & $0.46\pm0.04$ & $0.374\pm0.004$ \\
        & 5 & $0.236\pm0.005$ & $0.067\pm0.004$ & $0.768\pm0.081$ & $0.373\pm0.006$ \\
        \midrule
        \multirow{2}{3.6em}{Hopper}  & 0 & $0.141\pm0.005$ & $0.031\pm0.005$ & $0.428\pm0.034$ & $0.268\pm0.004$ \\
        & 5 & $0.145\pm0.003$ & $0.045\pm0.004$ & $0.616\pm0.05$ & $0.269\pm0.004$ \\
        \midrule
        \multirow{2}{3.6em}{Walker\\2d}  & 0 & $0.068\pm0.001$ & $0.011\pm0.003$ & $0.3\pm0.051$ & $0.159\pm0.002$ \\
        & 5 & $0.057\pm0.001$ & $0.017\pm0.002$ & $0.44\pm0.04$ & $0.159\pm0.002$ \\
        \bottomrule
    \end{tabular}
\end{sc}
\end{footnotesize} 
\end{center} 
\end{table}

\textbf{Dynamics Model Types.} Finally we compare the impact of the dynamics model type, in our case, either feed-forward(FF) or auto-regressive(AR). We use the same architecture as defined by  \citet{zhang2021autoregressive}. In \cref{table:gym-mujoco-dynamics-type} and \cref{table:maze-dynamics-type} (Appendix), we do not observe significant evidence in favor of the AR model with respect to OPCC performance. There is a marginal, but no statistical significant reduction in \emph{AURCC} in some cases. We do see an increase in coverage resolution ($CR_k$) for the gym-mujoco environments when using the AR model, while it remains the same for the maze2d environments. This may be due to the additional uncertainty propagation that can occur during auto-regressive inference of each dimension, especially in the higher-dimensional gym-mujoco environments. 
%
Currently our results do not suggest that the extra computational cost of the AR model compared to FF is worthwhile with respect to uncertainty quantification as measured via OPCC. This may be due to the environments not needed to represent multi-modal output distribution, which is where the AR model could have a distinct advantage. 

\textbf{Determinism.} Our baseline model is a deterministic version of the stochastic model defined in \citet{chua2018deep}, trained via regression loss. A classic improvement is to induce stochasticity into the model by learning a normal distribution over the next observation rather than a point estimate. We experimented with this modification and provide results in the Appendix (\cref{table:gym-mujoco-deterministic,table:maze-deterministic}).
Though, limitations of deterministic models are well-understood for stochastic environments, it turns out we don't gain significantly with stochastic models in our pilot run. This is possibly due to deterministic nature of maze environments and low stochasticity in gym-mujoco case. 

\begin{table}[tb!]
\caption{Evaluation metrics for  \emph{dynamics-type} comparison in \emph{gym-mujoco} environments comprising of mean and confidence interval(95$\%$) estimates over 50 samples belonging to 5(seeds) dynamics models for each of the 5 datasets.} 
\label{table:gym-mujoco-dynamics-type}
\begin{center} 
\begin{footnotesize} 
\begin{sc}
    \begin{tabular}{|l | c | c | c| c | c | c |}
        \toprule
        Env. & dynamics-type & AURCC$(\downarrow)$ & RPP$(\downarrow)$ & $CR_K(\uparrow)$ &  loss$(\downarrow)$ \\
        \midrule
        \multirow{2}{3.6em}{Half\\Cheetah}  & autoregressive & $0.249\pm0.008$ & $0.068\pm0.006$ & $0.736\pm0.087$ & $0.379\pm0.003$ \\
        & feed-forward & $0.219\pm0.005$ & $0.044\pm0.005$ & $0.46\pm0.04$ & $0.374\pm0.004$ \\
        \midrule
        \multirow{2}{3.6em}{Hopper}  & autoregressive & $0.139\pm0.007$ & $0.034\pm0.005$ & $0.48\pm0.042$ & $0.268\pm0.007$ \\
        & feed-forward & $0.141\pm0.005$ & $0.031\pm0.005$ & $0.428\pm0.034$ & $0.268\pm0.004$ \\
        \midrule
        \multirow{2}{3.6em}{Walker\\2d}  & autoregressive & $0.065\pm0.002$ & $0.013\pm0.002$ & $0.356\pm0.046$ & $0.158\pm0.002$ \\
        & feed-forward & $0.068\pm0.001$ & $0.011\pm0.003$ & $0.3\pm0.051$ & $0.159\pm0.002$ \\
        \bottomrule
    \end{tabular}
\end{sc}
\end{footnotesize} 
\end{center} 
\end{table}

\vspace{-0.2cm}
\section{Summary}
Properly quantifying uncertainty of complex models is a major open problem of practical significance in machine learning. Despite this fact, only a small fraction of the work in machine learning attempts to address this problem. Further, in areas such as offline RL, where methods for addressing uncertainty are developed, there is very little direct evaluation of uncertainty quantification. In recent years, there has been impressive progress on out-of-distribution detection for image classification, where quantifying uncertainty is a core problem. This has been largely driven by the availability of benchmarks that lower the overhead for conducting research and comparing methods. Currently, there is a lack of such benchmarks for sequential decision-making. The OPCC problem is a relatively simple problem to state, yet is rich enough to capture the essence of uncertainty quantification for sequential decision making. We hope that the OPCC benchmarks will inspire other researchers to develop new ideas for uncertainty quantification. Indeed, our pilot experiments show there is significant room to improve and that our understanding of current mechanisms is incomplete. Finally, we hope that this initial benchmark and baseline contribution is only the initial seed for the community at large to contribute to as progress is made. 

\bibliography{tmlr}

\begin{thebibliography}{56}
\providecommand{\natexlab}[1]{#1}
\providecommand{\url}[1]{\texttt{#1}}
\expandafter\ifx\csname urlstyle\endcsname\relax
  \providecommand{\doi}[1]{doi: #1}\else
  \providecommand{\doi}{doi: \begingroup \urlstyle{rm}\Url}\fi

\bibitem[Argenson \& Dulac-Arnold(2020)Argenson and
  Dulac-Arnold]{argenson2020model}
Arthur Argenson and Gabriel Dulac-Arnold.
\newblock Model-based offline planning.
\newblock \emph{arXiv preprint arXiv:2008.05556}, 2020.

\bibitem[Breiman(1996)]{breiman1996}
Leo Breiman.
\newblock Bagging predictors.
\newblock \emph{Machine learning}, 24\penalty0 (2):\penalty0 123--140, 1996.

\bibitem[Brockman et~al.(2016)Brockman, Cheung, Pettersson, Schneider,
  Schulman, Tang, and Zaremba]{brockman2016openai}
Greg Brockman, Vicki Cheung, Ludwig Pettersson, Jonas Schneider, John Schulman,
  Jie Tang, and Wojciech Zaremba.
\newblock Openai gym.
\newblock \emph{arXiv preprint arXiv:1606.01540}, 2016.

\bibitem[Buckman et~al.(2020)Buckman, Gelada, and
  Bellemare]{buckman2020importance}
Jacob Buckman, Carles Gelada, and Marc~G Bellemare.
\newblock The importance of pessimism in fixed-dataset policy optimization.
\newblock \emph{arXiv preprint arXiv:2009.06799}, 2020.

\bibitem[Chua et~al.(2018)Chua, Calandra, McAllister, and Levine]{chua2018deep}
Kurtland Chua, Roberto Calandra, Rowan McAllister, and Sergey Levine.
\newblock Deep reinforcement learning in a handful of trials using
  probabilistic dynamics models.
\newblock \emph{arXiv preprint arXiv:1805.12114}, 2018.

\bibitem[Clavera et~al.(2018)Clavera, Rothfuss, Schulman, Fujita, Asfour, and
  Abbeel]{clavera2018model}
Ignasi Clavera, Jonas Rothfuss, John Schulman, Yasuhiro Fujita, Tamim Asfour,
  and Pieter Abbeel.
\newblock Model-based reinforcement learning via meta-policy optimization.
\newblock In \emph{Conference on Robot Learning}, pp.\  617--629. PMLR, 2018.

\bibitem[Condessa et~al.(2017)Condessa, Bioucas-Dias, and
  Kova{\v{c}}evi{\'c}]{condessa2017performance}
Filipe Condessa, Jos{\'e} Bioucas-Dias, and Jelena Kova{\v{c}}evi{\'c}.
\newblock Performance measures for classification systems with rejection.
\newblock \emph{Pattern Recognition}, 63:\penalty0 437--450, 2017.

\bibitem[Dai et~al.(2020)Dai, Nachum, Chow, Li, Szepesv{\'a}ri, and
  Schuurmans]{dai2020coindice}
Bo~Dai, Ofir Nachum, Yinlam Chow, Lihong Li, Csaba Szepesv{\'a}ri, and Dale
  Schuurmans.
\newblock Coindice: Off-policy confidence interval estimation.
\newblock \emph{arXiv preprint arXiv:2010.11652}, 2020.

\bibitem[Deisenroth \& Rasmussen(2011)Deisenroth and
  Rasmussen]{deisenroth2011pilco}
Marc Deisenroth and Carl~E Rasmussen.
\newblock Pilco: A model-based and data-efficient approach to policy search.
\newblock In \emph{Proceedings of the 28th International Conference on machine
  learning (ICML-11)}, pp.\  465--472. Citeseer, 2011.

\bibitem[Dietterich(2000)]{dietterich2000ensemble}
Thomas~G Dietterich.
\newblock Ensemble methods in machine learning.
\newblock In \emph{International workshop on multiple classifier systems}, pp.\
   1--15. Springer, 2000.

\bibitem[Efron(1987)]{efron1987better}
Bradley Efron.
\newblock Better bootstrap confidence intervals.
\newblock \emph{Journal of the American statistical Association}, 82\penalty0
  (397):\penalty0 171--185, 1987.

\bibitem[Efron \& Tibshirani(1994)Efron and Tibshirani]{efron1994}
Bradley Efron and Robert~J Tibshirani.
\newblock \emph{An introduction to the bootstrap}.
\newblock CRC press, 1994.

\bibitem[El-Yaniv et~al.(2010)]{el-yaniv2010}
Ran El-Yaniv et~al.
\newblock On the foundations of noise-free selective classification.
\newblock \emph{Journal of Machine Learning Research}, 11\penalty0 (5), 2010.

\bibitem[Ernst et~al.(2005)Ernst, Geurts, and Wehenkel]{ernst2005tree}
Damien Ernst, Pierre Geurts, and Louis Wehenkel.
\newblock Tree-based batch mode reinforcement learning.
\newblock \emph{Journal of Machine Learning Research}, 6:\penalty0 503--556,
  2005.

\bibitem[Fu et~al.(2020)Fu, Kumar, Nachum, Tucker, and Levine]{fu2020d4rl}
Justin Fu, Aviral Kumar, Ofir Nachum, George Tucker, and Sergey Levine.
\newblock D4rl: Datasets for deep data-driven reinforcement learning.
\newblock \emph{arXiv preprint arXiv:2004.07219}, 2020.

\bibitem[Fu et~al.(2021)Fu, Norouzi, Nachum, Tucker, Wang, Novikov, Yang,
  Zhang, Chen, Kumar, et~al.]{fu2021benchmarks}
Justin Fu, Mohammad Norouzi, Ofir Nachum, George Tucker, Ziyu Wang, Alexander
  Novikov, Mengjiao Yang, Michael~R Zhang, Yutian Chen, Aviral Kumar, et~al.
\newblock Benchmarks for deep off-policy evaluation.
\newblock \emph{arXiv preprint arXiv:2103.16596}, 2021.

\bibitem[Fujimoto \& Gu(2021)Fujimoto and Gu]{fujimoto2021minimalist}
Scott Fujimoto and Shixiang~Shane Gu.
\newblock A minimalist approach to offline reinforcement learning.
\newblock \emph{arXiv preprint arXiv:2106.06860}, 2021.

\bibitem[Gal et~al.(2016)Gal, McAllister, and Rasmussen]{gal2016improving}
Yarin Gal, Rowan McAllister, and Carl~Edward Rasmussen.
\newblock Improving pilco with bayesian neural network dynamics models.
\newblock In \emph{Data-Efficient Machine Learning workshop, ICML}, volume~4,
  pp.\ ~25, 2016.

\bibitem[Geifman \& El-Yaniv(2017)Geifman and El-Yaniv]{geifman2017}
Yonatan Geifman and Ran El-Yaniv.
\newblock Selective classification for deep neural networks.
\newblock In \emph{Proceedings of the 31st International Conference on Neural
  Information Processing Systems}, pp.\  4885–4894, Red Hook, NY, USA, 2017.
  Curran Associates Inc.
\newblock ISBN 9781510860964.

\bibitem[Geifman \& El-Yaniv(2019)Geifman and El-Yaniv]{geifman2019}
Yonatan Geifman and Ran El-Yaniv.
\newblock Selectivenet: A deep neural network with an integrated reject option.
\newblock In \emph{International Conference on Machine Learning}, pp.\
  2151--2159. PMLR, 2019.

\bibitem[Hanna et~al.(2017)Hanna, Stone, and Niekum]{hanna2017bootstrapping}
Josiah~P Hanna, Peter Stone, and Scott Niekum.
\newblock Bootstrapping with models: Confidence intervals for off-policy
  evaluation.
\newblock In \emph{Thirty-First AAAI Conference on Artificial Intelligence},
  2017.

\bibitem[Hao et~al.(2021)Hao, Ji, Duan, Lu, Szepesv{\'a}ri, and
  Wang]{hao2021bootstrapping}
Botao Hao, Xiang Ji, Yaqi Duan, Hao Lu, Csaba Szepesv{\'a}ri, and Mengdi Wang.
\newblock Bootstrapping statistical inference for off-policy evaluation.
\newblock \emph{arXiv preprint arXiv:2102.03607}, 2021.

\bibitem[Hendrickx et~al.(2021)Hendrickx, Perini, Van~der Plas, Meert, and
  Davis]{hendrickx2021}
Kilian Hendrickx, Lorenzo Perini, Dries Van~der Plas, Wannes Meert, and Jesse
  Davis.
\newblock Machine learning with a reject option: A survey.
\newblock \emph{arXiv preprint arXiv:2107.11277}, 2021.

\bibitem[Janner et~al.(2019)Janner, Fu, Zhang, and Levine]{janner2019trust}
Michael Janner, Justin Fu, Marvin Zhang, and Sergey Levine.
\newblock When to trust your model: Model-based policy optimization.
\newblock \emph{arXiv preprint arXiv:1906.08253}, 2019.

\bibitem[Jin et~al.(2021{\natexlab{a}})Jin, Yang, and Wang]{jin2021pessimism}
Ying Jin, Zhuoran Yang, and Zhaoran Wang.
\newblock Is pessimism provably efficient for offline rl?
\newblock In \emph{International Conference on Machine Learning}, pp.\
  5084--5096. PMLR, 2021{\natexlab{a}}.

\bibitem[Jin et~al.(2021{\natexlab{b}})Jin, Zhang, Qin, Zhang, Yuan, Li, and
  Liu]{jin2021supervised}
Yue Jin, Yue Zhang, Tao Qin, Xudong Zhang, Jian Yuan, Houqiang Li, and Tie-Yan
  Liu.
\newblock Supervised off-policy ranking.
\newblock \emph{arXiv preprint arXiv:2107.01360}, 2021{\natexlab{b}}.

\bibitem[Kidambi et~al.(2020)Kidambi, Rajeswaran, Netrapalli, and
  Joachims]{kidambi2020morel}
Rahul Kidambi, Aravind Rajeswaran, Praneeth Netrapalli, and Thorsten Joachims.
\newblock Morel: Model-based offline reinforcement learning.
\newblock \emph{arXiv preprint arXiv:2005.05951}, 2020.

\bibitem[Kostrikov \& Nachum(2020)Kostrikov and
  Nachum]{kostrikov2020statistical}
Ilya Kostrikov and Ofir Nachum.
\newblock Statistical bootstrapping for uncertainty estimation in off-policy
  evaluation.
\newblock \emph{arXiv preprint arXiv:2007.13609}, 2020.

\bibitem[Kostrikov et~al.(2021)Kostrikov, Fergus, Tompson, and
  Nachum]{kostrikov2021offline}
Ilya Kostrikov, Rob Fergus, Jonathan Tompson, and Ofir Nachum.
\newblock Offline reinforcement learning with fisher divergence critic
  regularization.
\newblock In \emph{International Conference on Machine Learning}, pp.\
  5774--5783. PMLR, 2021.

\bibitem[Kumar et~al.(2019)Kumar, Fu, Tucker, and Levine]{kumar2019stabilizing}
Aviral Kumar, Justin Fu, George Tucker, and Sergey Levine.
\newblock Stabilizing off-policy q-learning via bootstrapping error reduction.
\newblock \emph{arXiv preprint arXiv:1906.00949}, 2019.

\bibitem[Kumar et~al.(2020)Kumar, Zhou, Tucker, and
  Levine]{kumar2020conservative}
Aviral Kumar, Aurick Zhou, George Tucker, and Sergey Levine.
\newblock Conservative q-learning for offline reinforcement learning.
\newblock \emph{arXiv preprint arXiv:2006.04779}, 2020.

\bibitem[Kurutach et~al.(2018)Kurutach, Clavera, Duan, Tamar, and
  Abbeel]{kurutach2018model}
Thanard Kurutach, Ignasi Clavera, Yan Duan, Aviv Tamar, and Pieter Abbeel.
\newblock Model-ensemble trust-region policy optimization.
\newblock \emph{arXiv preprint arXiv:1802.10592}, 2018.

\bibitem[Levine et~al.(2020)Levine, Kumar, Tucker, and Fu]{levine2020offline}
Sergey Levine, Aviral Kumar, George Tucker, and Justin Fu.
\newblock Offline reinforcement learning: Tutorial, review, and perspectives on
  open problems.
\newblock \emph{arXiv preprint arXiv:2005.01643}, 2020.

\bibitem[Loh(1987)]{loh1987}
Wei-Yin Loh.
\newblock Calibrating confidence coefficients.
\newblock \emph{Journal of the American Statistical Association}, 82\penalty0
  (397):\penalty0 155--162, 1987.

\bibitem[Massart(2007)]{massart2007concentration}
Pascal Massart.
\newblock \emph{Concentration inequalities and model selection: Ecole d'Et{\'e}
  de Probabilit{\'e}s de Saint-Flour XXXIII-2003}.
\newblock Springer, 2007.

\bibitem[Mnih et~al.(2013)Mnih, Kavukcuoglu, Silver, Graves, Antonoglou,
  Wierstra, and Riedmiller]{mnih2013playing}
Volodymyr Mnih, Koray Kavukcuoglu, David Silver, Alex Graves, Ioannis
  Antonoglou, Daan Wierstra, and Martin Riedmiller.
\newblock Playing atari with deep reinforcement learning.
\newblock \emph{arXiv preprint arXiv:1312.5602}, 2013.

\bibitem[Nachum et~al.(2019)Nachum, Chow, Dai, and Li]{nachum2019dualdice}
Ofir Nachum, Yinlam Chow, Bo~Dai, and Lihong Li.
\newblock Dualdice: Behavior-agnostic estimation of discounted stationary
  distribution corrections.
\newblock \emph{arXiv preprint arXiv:1906.04733}, 2019.

\bibitem[Naeini et~al.(2015)Naeini, Cooper, and Hauskrecht]{naeini2015}
Mahdi~Pakdaman Naeini, Gregory Cooper, and Milos Hauskrecht.
\newblock Obtaining well calibrated probabilities using bayesian binning.
\newblock In \emph{Twenty-Ninth AAAI Conference on Artificial Intelligence},
  2015.

\bibitem[Nagabandi et~al.(2020)Nagabandi, Konolige, Levine, and
  Kumar]{nagabandi2020deep}
Anusha Nagabandi, Kurt Konolige, Sergey Levine, and Vikash Kumar.
\newblock Deep dynamics models for learning dexterous manipulation.
\newblock In \emph{Conference on Robot Learning}, pp.\  1101--1112. PMLR, 2020.

\bibitem[Osband et~al.(2018)Osband, Aslanides, and
  Cassirer]{osband2018randomized}
Ian Osband, John Aslanides, and Albin Cassirer.
\newblock Randomized prior functions for deep reinforcement learning.
\newblock \emph{arXiv preprint arXiv:1806.03335}, 2018.

\bibitem[Peng et~al.(2019)Peng, Kumar, Zhang, and Levine]{peng2019advantage}
Xue~Bin Peng, Aviral Kumar, Grace Zhang, and Sergey Levine.
\newblock Advantage-weighted regression: Simple and scalable off-policy
  reinforcement learning.
\newblock \emph{arXiv preprint arXiv:1910.00177}, 2019.

\bibitem[Puterman(2014)]{puterman2014}
Martin~L Puterman.
\newblock \emph{Markov decision processes: discrete stochastic dynamic
  programming}.
\newblock John Wiley \& Sons, 2014.

\bibitem[Rafailov et~al.(2021)Rafailov, Yu, Rajeswaran, and
  Finn]{rafailov2021offline}
Rafael Rafailov, Tianhe Yu, Aravind Rajeswaran, and Chelsea Finn.
\newblock Offline reinforcement learning from images with latent space models.
\newblock In \emph{Learning for Dynamics and Control}, pp.\  1154--1168. PMLR,
  2021.

\bibitem[Rasmussen(2003)]{rasmussen2003gaussian}
Carl~Edward Rasmussen.
\newblock Gaussian processes in machine learning.
\newblock In \emph{Summer school on machine learning}, pp.\  63--71. Springer,
  2003.

\bibitem[Richards(2005)]{richards2005robust}
Arthur~George Richards.
\newblock \emph{Robust constrained model predictive control}.
\newblock PhD thesis, Massachusetts Institute of Technology, 2005.

\bibitem[Schrittwieser et~al.(2021)Schrittwieser, Hubert, Mandhane, Barekatain,
  Antonoglou, and Silver]{schrittwieser2021online}
Julian Schrittwieser, Thomas Hubert, Amol Mandhane, Mohammadamin Barekatain,
  Ioannis Antonoglou, and David Silver.
\newblock Online and offline reinforcement learning by planning with a learned
  model.
\newblock \emph{arXiv preprint arXiv:2104.06294}, 2021.

\bibitem[Schulman et~al.(2017)Schulman, Wolski, Dhariwal, Radford, and
  Klimov]{schulman2017}
John Schulman, Filip Wolski, Prafulla Dhariwal, Alec Radford, and Oleg Klimov.
\newblock Proximal policy optimization algorithms.
\newblock \emph{arXiv preprint arXiv:1707.06347}, 2017.

\bibitem[Shrestha et~al.(2021)Shrestha, Lee, Tadepalli, and Fern]{shrestha2021}
Aayam Shrestha, Stefan Lee, Prasad Tadepalli, and Alan Fern.
\newblock Deepaveragers: Offline reinforcement learning by solving derived
  non-parametric mdps.
\newblock In \emph{International Conference on Learning Representations}, 2021.

\bibitem[Sonabend-W et~al.(2020)Sonabend-W, Lu, Celi, Cai, and
  Szolovits]{sonabend2020expert}
Aaron Sonabend-W, Junwei Lu, Leo~A Celi, Tianxi Cai, and Peter Szolovits.
\newblock Expert-supervised reinforcement learning for offline policy learning
  and evaluation.
\newblock \emph{arXiv preprint arXiv:2006.13189}, 2020.

\bibitem[Strehl \& Littman(2008)Strehl and Littman]{strehl2008analysis}
Alexander~L Strehl and Michael~L Littman.
\newblock An analysis of model-based interval estimation for markov decision
  processes.
\newblock \emph{Journal of Computer and System Sciences}, 74\penalty0
  (8):\penalty0 1309--1331, 2008.

\bibitem[Thomas et~al.(2015)Thomas, Theocharous, and
  Ghavamzadeh]{thomas2015high}
Philip Thomas, Georgios Theocharous, and Mohammad Ghavamzadeh.
\newblock High-confidence off-policy evaluation.
\newblock In \emph{Proceedings of the AAAI Conference on Artificial
  Intelligence}, volume~29, 2015.

\bibitem[Xin et~al.(2021)Xin, Tang, Yu, and Lin]{xin2021}
Ji~Xin, Raphael Tang, Yaoliang Yu, and Jimmy Lin.
\newblock The art of abstention: Selective prediction and error regularization
  for natural language processing.
\newblock In \emph{Proceedings of the 59th Annual Meeting of the Association
  for Computational Linguistics and the 11th International Joint Conference on
  Natural Language Processing (Volume 1: Long Papers)}, pp.\  1040--1051, 2021.

\bibitem[Yu et~al.(2020)Yu, Thomas, Yu, Ermon, Zou, Levine, Finn, and
  Ma]{yu2020mopo}
Tianhe Yu, Garrett Thomas, Lantao Yu, Stefano Ermon, James Zou, Sergey Levine,
  Chelsea Finn, and Tengyu Ma.
\newblock Mopo: Model-based offline policy optimization.
\newblock \emph{arXiv preprint arXiv:2005.13239}, 2020.

\bibitem[Yu et~al.(2021)Yu, Kumar, Rafailov, Rajeswaran, Levine, and
  Finn]{yu2021combo}
Tianhe Yu, Aviral Kumar, Rafael Rafailov, Aravind Rajeswaran, Sergey Levine,
  and Chelsea Finn.
\newblock Combo: Conservative offline model-based policy optimization.
\newblock \emph{arXiv preprint arXiv:2102.08363}, 2021.

\bibitem[Zhang et~al.(2021)Zhang, Paine, Nachum, Paduraru, Tucker, Wang, and
  Norouzi]{zhang2021autoregressive}
Michael~R Zhang, Tom~Le Paine, Ofir Nachum, Cosmin Paduraru, George Tucker,
  Ziyu Wang, and Mohammad Norouzi.
\newblock Autoregressive dynamics models for offline policy evaluation and
  optimization.
\newblock \emph{arXiv preprint arXiv:2104.13877}, 2021.

\bibitem[Zhang et~al.(2020)Zhang, Liu, and Whiteson]{zhang2020gradientdice}
Shangtong Zhang, Bo~Liu, and Shimon Whiteson.
\newblock Gradientdice: Rethinking generalized offline estimation of stationary
  values.
\newblock In \emph{International Conference on Machine Learning}, pp.\
  11194--11203. PMLR, 2020.

\end{thebibliography}
\bibliographystyle{tmlr}

\newpage
\appendix
\section{Appendix - OPCC Benchmark Summary}
\label{sec:opcc-benchmark-summary-appendix}

In the following, we share a snapshot of OPCC benchmark components.
\begin{table}[h!]
\caption{ Information about OPCC Benchmark comprising of environment details , datasets and queries.} 
\label{table:opcc-benchmark}
\begin{center} 
\begin{footnotesize} 
\begin{sc} 
\begin{tabular}{|p{0.18\textwidth}| p{0.1\textwidth} |p{0.1\textwidth} | p{0.1\textwidth}| p{0.3\textwidth}| p{0.1\textwidth}|}
\toprule
Env. & Observation-size & Action-Dimensions & Max. Env. Steps & Dataset-name & Query-count\\
\midrule
maze2d-open-v0 &4 &2 & 150 &1m & $1500$ \\ 
\midrule 
maze2d-medium-v1 & 4&2 & 600  &1m & $1500$ \\ 
\midrule 
maze2d-umaze-v1 & 4&2  & 300 & 1m & $1500$ \\ 
\midrule 
maze2d-umaze-v1 &4 &2&  800 & 1m & \cellcolor{yellow}$121$ \\ 
\midrule
HalfCheetah-v2 & 17 & 6 & 1000 &random, expert, medium, medium-replay, medium-expert	& 1500 \\
\midrule
Hopper-v2	&11&3 & 1000 &random, expert, medium, medium-replay, medium-expert	& 1500 \\
\midrule
Walker2d-v2	& 17& 6 & 1000 & random, expert, medium, medium-replay, medium-expert &	1500 \\
\bottomrule 
\end{tabular}
\end{sc}
\end{footnotesize} 
\end{center} 
\end{table}

\begin{table}[h!]
\caption{Performance of policies used in PCQs. These policies are trained using PPO \citep{schulman2017} over the original environment task and hand-picked at different performance levels. We report mean and standard deviation of policy performance over 20 episodes.} 
\label{table:opcc-benchmark-policies}
\begin{center} 
\begin{footnotesize} 
\begin{sc} 
\begin{tabular}{|l | c | c | c | c|}
\toprule
Env. & policy-1 & policy-2 & policy-3 & policy-4 \\
\midrule
maze2d-open-v0	& $122 \pm 10$ & $104 \pm 22$ & $18 \pm 14$ & $4 \pm 8$ \\
\midrule
maze2d-umaze-v1 & $245 \pm 272$ & $203 \pm 252$  & $256 \pm 260$ & $258 \pm 262$\\
\midrule
maze2d-medium-v1 & $235 \pm 35$ & $197 \pm 58$ & $23 \pm 73$ & $3 \pm 9$\\
\midrule
maze2d-large-v1	& $231 \pm 268$ & $160 \pm 201$ & $50 \pm 76$ & $9 \pm 9$\\
\midrule
HalfCheetah-v2	& $1168 \pm 80$ & $1044 \pm 112$ & $785 \pm 303$ & $94 \pm 40$\\
\midrule
Hopper-v2	& $1195 \pm 794$ & $1466 \pm 487$ & $1832 \pm 560$ & $236 \pm 1$\\
\midrule
Walker2d-v2	& $2506 \pm 698$ & $811 \pm 321$ & $387 \pm 42$ & $162 \pm 102$ \\
\bottomrule 
\end{tabular}
\end{sc}
\end{footnotesize} 
\end{center} 
\end{table}

\begin{figure}[h!]
    \centering
    \includegraphics[scale=0.7]{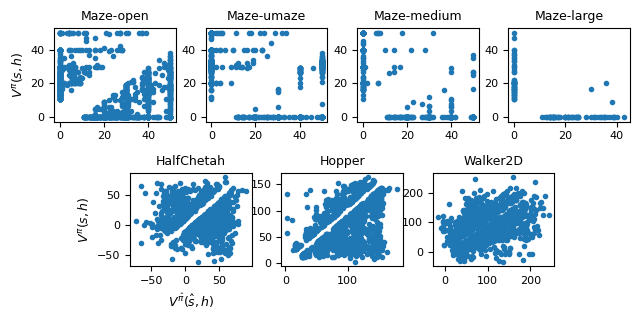}
    \caption{Scatter plot of the PCQ for each benchmark environment. For each PCQ $(s,\pi,\hs,\hpi,h)$, we plot $V^{\pi}(s,h)$ vs. $V^{\hpi}(\hs,h)$.}
    \label{fig:query_viz_all}
\end{figure}

\newpage
\section{Appendix - Evaluation Metrics \& Selective-Risk Coverage Curves for Ablations}

In the following sub-sections, we share OPCC metrics for various ablations over our baseline. In each table-cell, we show mean and confidence interval at \emph{95$\%$ confidence level} for corresponding metrics, estimated by evaluating 5 dynamics runs over each dataset of the corresponding environment.

\subsection{Randomized Constant Priors}
\begin{figure}[h!]
\centering
\includegraphics[scale=0.7]{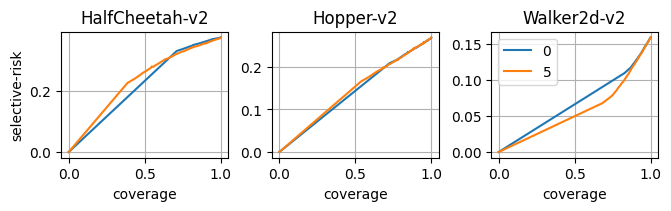}
\caption{Selective-risk coverage curves for  \emph{prior-scale} in \emph{gym-mujoco} environments.} 
\label{fig:gym-mujoco-prior-scale}
\end{figure}

\begin{figure}[h!]
\centering
\includegraphics[scale=0.7]{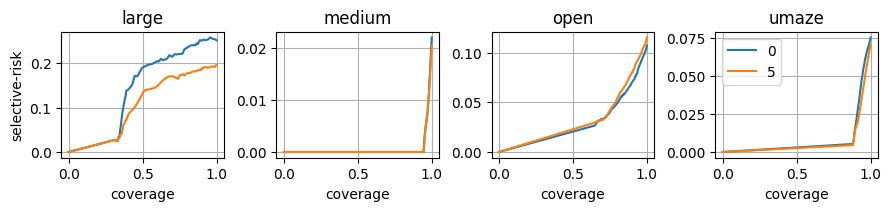}
\caption{Selective-risk coverage curves for  \emph{prior-scale} in \emph{maze} environments } 
\label{fig:maze-prior-scale}
\end{figure}

 Figures \ref{fig:gym-mujoco-prior-scale},\ref{fig:maze-prior-scale} and Tables \ref{table:gym-mujoco-prior-scale},\ref{table:maze-prior-scale}  shows impact of adding randomized constant priors to our baseline ensemble models. Output of prior models is scaled by \emph{Prior-Scale} before adding to dynamic model. \emph{Prior-Scale} of 0 implies randomized prior was not added. We observe significant performance gain only in Large-Maze environment.

\begin{table}[h!]
\caption{Evaluation metrics for  \emph{prior-scale} comparison in \emph{maze} environments.} 
\label{table:maze-prior-scale}
\begin{center} 
\begin{footnotesize} 
\begin{sc}
    \begin{tabular}{|l | c | c | c| c | c | c |}
        \toprule
        Env. & prior-scale & AURCC$(\downarrow)$ & RPP$(\downarrow)$ & $CR_K(\uparrow)$ &  loss$(\downarrow)$ \\
        \midrule
        \multirow{2}{3.6em}{large}  & 0 & $0.14\pm0.015$ & $0.062\pm0.004$ & $0.82\pm0.035$ & $0.251\pm0.029$ \\
        & 5 & $0.104\pm0.017$ & $0.051\pm0.012$ & $0.82\pm0.035$ & $0.197\pm0.017$ \\
        \midrule
        \multirow{2}{3.6em}{medium}  & 0 & $0.001\pm(<0.001)$ & $0.0\pm(<0.001)$ & $0.2\pm(<0.001)$ & $0.022\pm0.001$ \\
        & 5 & $0.0\pm(<0.001)$ & $0.0\pm(<0.001)$ & $0.2\pm(<0.001)$ & $0.02\pm0.003$ \\
        \midrule
        \multirow{2}{3.6em}{open}  & 0 & $0.029\pm0.001$ & $0.012\pm0.001$ & $0.5\pm(<0.001)$ & $0.107\pm0.005$ \\
        & 5 & $0.032\pm0.001$ & $0.012\pm(<0.001)$ & $0.5\pm(<0.001)$ & $0.115\pm0.008$ \\
        \midrule
        \multirow{2}{3.6em}{umaze}  & 0 & $0.008\pm0.001$ & $0.002\pm(<0.001)$ & $0.3\pm(<0.001)$ & $0.075\pm0.003$ \\
        & 5 & $0.006\pm0.001$ & $0.002\pm(<0.001)$ & $0.3\pm(<0.001)$ & $0.071\pm0.002$ \\
        \bottomrule
    \end{tabular}
\end{sc}
\end{footnotesize} 
\end{center} 
\end{table}


\newpage
\subsection{Deterministic Model}
\begin{figure}[h!]
\centering
\includegraphics[scale=0.63]{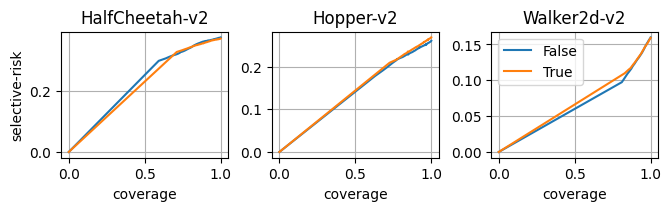}
\caption{Selective-risk coverage curves for  \emph{deterministic} in \emph{gym-mujoco} environments } 
\label{fig:gym-mujoco-deterministic}
\end{figure}

\begin{figure}[h!]
\centering
\includegraphics[scale=0.63]{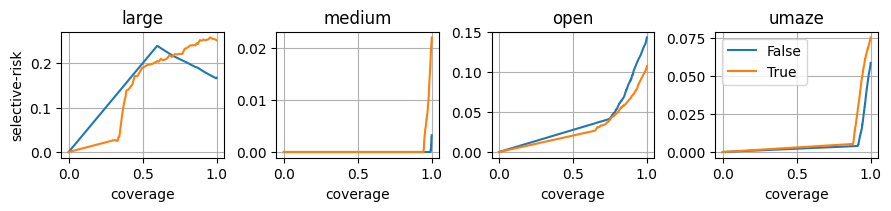}
\caption{Selective-risk coverage curves for  \emph{deterministic} in \emph{maze} environments } 
\label{fig:maze-deterministic}
\end{figure}

In Tables \ref{table:gym-mujoco-deterministic}, \ref{table:maze-deterministic} and Figures \ref{fig:gym-mujoco-deterministic},\ref{fig:maze-deterministic}, we share the choice of having a \emph{deterministic(True)} versus \emph{stochastic(False)} dynamics model. Though, the mean of stochastic model is lower, it's not significant as it's within confidence interval of the deterministic model.

\begin{table}[h!]
\caption{Evaluation metrics for  \emph{deterministic} model comparison in \emph{gym-mujoco} environments. } 
\label{table:gym-mujoco-deterministic}
\begin{center} 
\begin{footnotesize} 
\begin{sc}
    \begin{tabular}{|l | c | c | c| c | c | c |}
        \toprule
        Env. & deterministic & AURCC$(\downarrow)$ & RPP$(\downarrow)$ & $CR_K(\uparrow)$ &  loss$(\downarrow)$ \\
        \midrule
        \multirow{2}{3.6em}{Half\\Cheetah}  & False & $0.229\pm0.007$ & $0.054\pm0.006$ & $0.568\pm0.057$ & $0.377\pm0.005$ \\
        & True & $0.219\pm0.005$ & $0.044\pm0.005$ & $0.46\pm0.04$ & $0.374\pm0.004$ \\
        \midrule
        \multirow{2}{3.6em}{Hopper}  & False & $0.138\pm0.002$ & $0.039\pm0.002$ & $0.524\pm0.039$ & $0.26\pm0.003$ \\
        & True & $0.141\pm0.005$ & $0.031\pm0.005$ & $0.428\pm0.034$ & $0.268\pm0.004$ \\
        \midrule
        \multirow{2}{3.6em}{Walker\\2d}  & False & $0.064\pm(<0.001)$ & $0.012\pm0.001$ & $0.328\pm0.024$ & $0.16\pm0.002$ \\
        & True & $0.068\pm0.001$ & $0.011\pm0.003$ & $0.3\pm0.051$ & $0.159\pm0.002$ \\
        \bottomrule
    \end{tabular}
\end{sc}
\end{footnotesize} 
\end{center} 
\end{table}

\begin{table}[h!]
\caption{Evaluation metrics for  \emph{deterministic} model comparison in \emph{maze} environments.} 
\label{table:maze-deterministic}
\begin{center} 
\begin{footnotesize} 
\begin{sc}
    \begin{tabular}{|l | c | c | c| c | c | c |}
        \toprule
        Env. & deterministic & AURCC$(\downarrow)$ & RPP$(\downarrow)$ & $CR_K(\uparrow)$ &  loss$(\downarrow)$ \\
        \midrule
        \multirow{2}{3.6em}{large}  & False & $0.152\pm0.01$ & $0.058\pm0.007$ & $0.54\pm0.043$ & $0.167\pm0.003$ \\
        & True & $0.14\pm0.015$ & $0.062\pm0.004$ & $0.82\pm0.035$ & $0.251\pm0.029$ \\
        \midrule
        \multirow{2}{3.6em}{medium}  & False & $0.0\pm(<0.001)$ & $0.0\pm(<0.001)$ & $0.2\pm(<0.001)$ & $0.003\pm(<0.001)$ \\
        & True & $0.001\pm(<0.001)$ & $0.0\pm(<0.001)$ & $0.2\pm(<0.001)$ & $0.022\pm0.001$ \\
        \midrule
        \multirow{2}{3.6em}{open}  & False & $0.037\pm0.001$ & $0.01\pm(<0.001)$ & $0.4\pm(<0.001)$ & $0.143\pm0.005$ \\
        & True & $0.029\pm0.001$ & $0.012\pm0.001$ & $0.5\pm(<0.001)$ & $0.107\pm0.005$ \\
        \midrule
        \multirow{2}{3.6em}{umaze}  & False & $0.004\pm(<0.001)$ & $0.001\pm(<0.001)$ & $0.2\pm(<0.001)$ & $0.059\pm0.004$ \\
        & True & $0.008\pm0.001$ & $0.002\pm(<0.001)$ & $0.3\pm(<0.001)$ & $0.075\pm0.003$ \\
        \bottomrule
    \end{tabular}
\end{sc}
\end{footnotesize} 
\end{center} 
\end{table}


\newpage
\subsection{Dynamics Type}
\begin{table}[h!]
\caption{Evaluation metrics for  \emph{dynamics-type} comparison in \emph{maze} environments } 
\label{table:maze-dynamics-type}
\begin{center} 
\begin{footnotesize} 
\begin{sc}
    \begin{tabular}{|l | c | c | c| c | c | c |}
        \toprule
        Env. & dynamics-type & AURCC$(\downarrow)$ & RPP$(\downarrow)$ & $CR_K(\uparrow)$ &  loss$(\downarrow)$ \\
        \midrule
        \multirow{2}{3.6em}{large}  & autoregressive & $0.131\pm0.017$ & $0.06\pm0.003$ & $0.8\pm(<0.001)$ & $0.233\pm0.044$ \\
        & feed-forward & $0.14\pm0.015$ & $0.062\pm0.004$ & $0.82\pm0.035$ & $0.251\pm0.029$ \\
        \midrule
        \multirow{2}{3.6em}{medium}  & autoregressive & $0.001\pm(<0.001)$ & $0.0\pm(<0.001)$ & $0.2\pm(<0.001)$ & $0.031\pm0.001$ \\
        & feed-forward & $0.001\pm(<0.001)$ & $0.0\pm(<0.001)$ & $0.2\pm(<0.001)$ & $0.022\pm0.001$ \\
        \midrule
        \multirow{2}{3.6em}{open}  & autoregressive & $0.034\pm0.003$ & $0.014\pm0.002$ & $0.5\pm(<0.001)$ & $0.123\pm0.006$ \\
        & feed-forward & $0.029\pm0.001$ & $0.012\pm0.001$ & $0.5\pm(<0.001)$ & $0.107\pm0.005$ \\
        \midrule
        \multirow{2}{3.6em}{umaze}  & autoregressive & $0.007\pm0.001$ & $0.002\pm(<0.001)$ & $0.3\pm(<0.001)$ & $0.07\pm0.002$ \\
        & feed-forward & $0.008\pm0.001$ & $0.002\pm(<0.001)$ & $0.3\pm(<0.001)$ & $0.075\pm0.003$ \\
        \bottomrule
    \end{tabular}
\end{sc}
\end{footnotesize} 
\end{center} 
\end{table}

\begin{figure}[h!]
\centering
\includegraphics[scale=0.7]{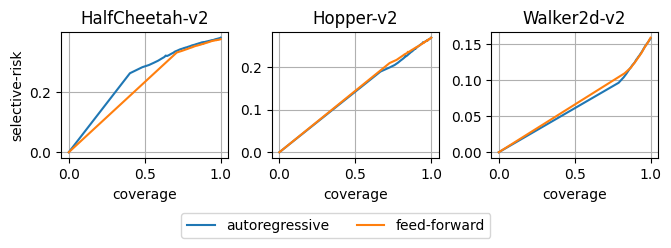}
\caption{Selective-risk coverage curves for  \emph{dynamics-type} in \emph{gym-mujoco} environments } 
\label{fig:gym-mujoco-dynamics-type}
\end{figure}
\begin{figure}[h!]
\centering
\includegraphics[scale=0.7]{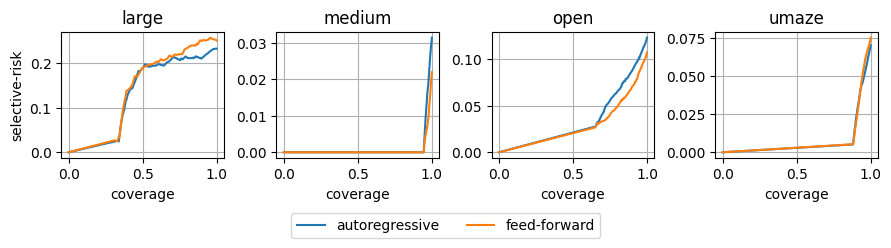}
\caption{Selective-risk coverage curves for  \emph{dynamics-type} in \emph{maze} environments } 
\label{fig:maze-dynamics-type}
\end{figure}

\newpage
\subsection{Normalization of input state-space}
\begin{figure}[h!]
\centering
\includegraphics[scale=0.6]{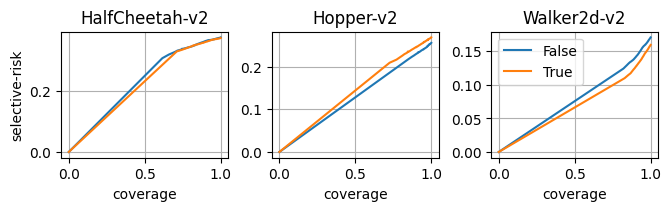}
\caption{Selective-risk coverage curves for  \emph{normalize} in \emph{gym-mujoco} environments. }
\label{fig:gym-mujoco-normalize}
\end{figure}

\begin{figure}[h!]
\centering
\includegraphics[scale=0.6]{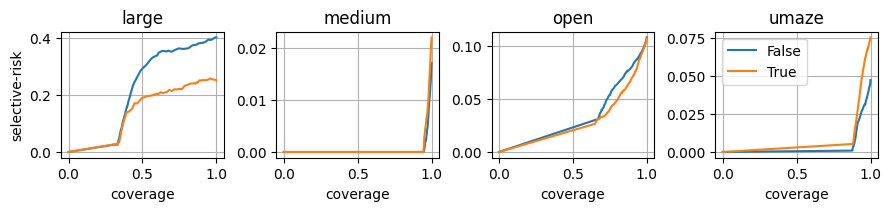}
\caption{Selective-risk coverage curves for  \emph{normalize} in \emph{maze} environments.} 
\label{fig:maze-normalize}
\end{figure}

In Tables \ref{table:gym-mujoco-normalize},\ref{table:maze-normalize} and Figures \ref{fig:gym-mujoco-normalize},\ref{fig:maze-normalize}, we investigate the imapct of learning dynamics with normalized state-space. Here, \emph{'True'} implies the dynamics was learned with normalized state-space and \emph{'False'} implies otherwise. There is marginal performance difference between either choice for MuJoco environments.  Maze2D environments show a mix of results with normalization benefiting Umaze and hurting Large-Maze. Performance of Medium-Maze and Open-Maze is not impacted significantly.

\begin{table}[h!]
\caption{Evaluation metrics for  \emph{normalize} comparison in \emph{gym-mujoco} environments } 
\label{table:gym-mujoco-normalize}
\begin{center} 
\begin{footnotesize} 
\begin{sc} 
\begin{tabular}{|l | c | c | c| c | c | c |}
\toprule
Env. & normalize & AURCC$(\downarrow)$ & RPP$(\downarrow)$ & $CR_K(\uparrow)$ &  loss$(\downarrow)$ \\
\midrule
\multirow{2}{3.6em}{Half\\Cheetah}  & False & $0.228\pm0.006$ & $0.053\pm0.007$ & $0.548\pm0.07$ & $0.376\pm0.003$ \\
& True & $0.219\pm0.005$ & $0.044\pm0.005$ & $0.46\pm0.04$ & $0.374\pm0.004$ \\
\midrule
\multirow{2}{3.6em}{Hopper}  & False & $0.128\pm0.002$ & $0.017\pm0.002$ & $0.3\pm0.025$ & $0.255\pm0.003$ \\
& True & $0.141\pm0.005$ & $0.031\pm0.005$ & $0.428\pm0.034$ & $0.268\pm0.004$ \\
\midrule
\multirow{2}{3.6em}{Walker\\2d}  & False & $0.078\pm0.007$ & $0.013\pm0.005$ & $0.316\pm0.073$ & $0.17\pm0.009$ \\
& True & $0.068\pm0.001$ & $0.011\pm0.003$ & $0.3\pm0.051$ & $0.159\pm0.002$ \\
\bottomrule
\end{tabular}
\end{sc}
\end{footnotesize} 
\end{center} 
\end{table}

\vspace{-0.1cm}
\begin{table}[h!]
\caption{Evaluation metrics for  \emph{normalize} comparison in \emph{maze} environments } 
\label{table:maze-normalize}
\begin{center} 
\begin{footnotesize} 
\begin{sc}
    \begin{tabular}{|l | c | c | c| c | c | c |}
        \toprule
        Env. & normalize & AURCC$(\downarrow)$ & RPP$(\downarrow)$ & $CR_K(\uparrow)$ &  loss$(\downarrow)$ \\
        \midrule
        \multirow{2}{3.6em}{large}  & False & $0.215\pm0.027$ & $0.067\pm0.007$ & $0.82\pm0.035$ & $0.402\pm0.037$ \\
        & True & $0.14\pm0.015$ & $0.062\pm0.004$ & $0.82\pm0.035$ & $0.251\pm0.029$ \\
        \midrule
        \multirow{2}{3.6em}{medium}  & False & $0.0\pm(<0.001)$ & $0.0\pm(<0.001)$ & $0.2\pm(<0.001)$ & $0.017\pm0.002$ \\
        & True & $0.001\pm(<0.001)$ & $0.0\pm(<0.001)$ & $0.2\pm(<0.001)$ & $0.022\pm0.001$ \\
        \midrule
        \multirow{2}{3.6em}{open}  & False & $0.033\pm0.001$ & $0.014\pm(<0.001)$ & $0.5\pm(<0.001)$ & $0.108\pm0.004$ \\
        & True & $0.029\pm0.001$ & $0.012\pm0.001$ & $0.5\pm(<0.001)$ & $0.107\pm0.005$ \\
        \midrule
        \multirow{2}{3.6em}{umaze}  & False & $0.003\pm(<0.001)$ & $0.002\pm(<0.001)$ & $0.3\pm(<0.001)$ & $0.047\pm0.004$ \\
        & True & $0.008\pm0.001$ & $0.002\pm(<0.001)$ & $0.3\pm(<0.001)$ & $0.075\pm0.003$ \\
        \bottomrule
    \end{tabular}
\end{sc}
\end{footnotesize} 
\end{center} 
\end{table}


\newpage
\subsection{Uncertainty Types}
In the following, \emph{'EV, PCI, U-PCI'} refer to Ensemble-Voting, Paired Confidence interval and Unpaired Confidence Interval, respectively.
\begin{table}[h!]
\caption{Evaluation metrics for  \emph{uncertainty-type} comparison in \emph{maze} environments } 
\label{table:maze-uncertainty-type}
\begin{center} 
\begin{footnotesize} 
\begin{sc}
    \begin{tabular}{|l | c | c | c| c | c | c |}
        \toprule
        Env. & uncertainty-type & AURCC$(\downarrow)$ & RPP$(\downarrow)$ & $CR_K(\uparrow)$ &  loss$(\downarrow)$ \\
        \midrule
        \multirow{3}{3.6em}{medium}  & ev & $0.001\pm(<0.001)$ & $0.0\pm(<0.001)$ & $0.2\pm(<0.001)$ & $0.022\pm0.001$ \\
        & pci & $0.001\pm(<0.001)$ & $0.0\pm(<0.001)$ & $0.2\pm(<0.001)$ & $0.009\pm0.001$ \\
        & u-pci & $0.001\pm(<0.001)$ & $0.0\pm(<0.001)$ & $0.2\pm(<0.001)$ & $0.009\pm0.001$ \\
        \midrule
        \multirow{3}{3.6em}{open}  & ev & $0.029\pm0.001$ & $0.012\pm0.001$ & $0.5\pm(<0.001)$ & $0.107\pm0.005$ \\
        & pci & $0.057\pm0.004$ & $0.012\pm0.001$ & $0.38\pm0.035$ & $0.168\pm0.008$ \\
        & u-pci & $0.05\pm0.002$ & $0.012\pm0.001$ & $0.4\pm(<0.001)$ & $0.168\pm0.008$ \\
        \midrule
        \multirow{3}{3.6em}{umaze}  & ev & $0.008\pm0.001$ & $0.002\pm(<0.001)$ & $0.3\pm(<0.001)$ & $0.075\pm0.003$ \\
        & pci & $0.035\pm0.002$ & $0.001\pm(<0.001)$ & $0.2\pm(<0.001)$ & $0.084\pm0.003$ \\
        & u-pci & $0.017\pm0.002$ & $0.001\pm(<0.001)$ & $0.2\pm(<0.001)$ & $0.084\pm0.003$ \\
        \bottomrule
    \end{tabular}
\end{sc}
\end{footnotesize} 
\end{center} 
\end{table}

\begin{figure}[h!]
\centering
\includegraphics[scale=0.7]{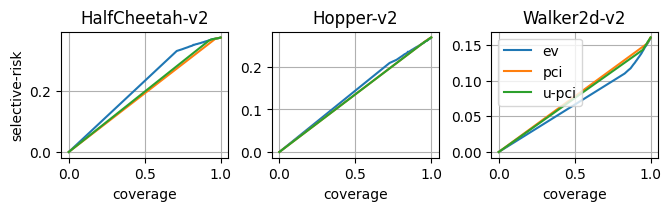}
\caption{Selective-risk coverage curves for  \emph{uncertainty-type} in \emph{gym-mujoco} environments.}  
\label{fig:gym-mujoco-uncertainty-type}
\end{figure}
\begin{figure}[h!]
\centering
\includegraphics[scale=0.7]{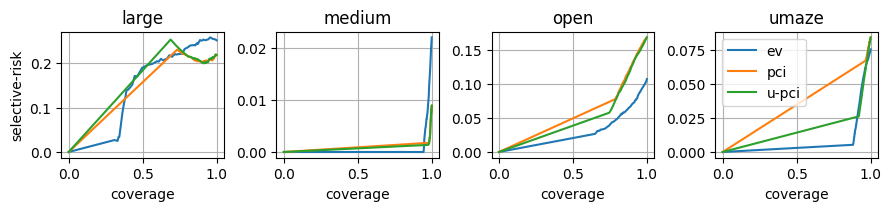}
\caption{Selective-risk coverage curves for  \emph{uncertainty-type} in \emph{maze} environments.} 
\label{fig:maze-uncertainty-type}
\end{figure}

\newpage
\subsection{Horizon}
\begin{table}[h!]
\caption{Evaluation metrics for  \emph{horizon} comparison in \emph{maze} environments } 
\label{table:maze-horizon}
\begin{center} 
\begin{footnotesize} 
\begin{sc}
    \begin{tabular}{|l | c | c | c| c | c | c |}
        \toprule
        Env. & horizon & AURCC$(\downarrow)$ & RPP$(\downarrow)$ & $CR_K(\uparrow)$ &  loss$(\downarrow)$ \\
        \toprule
        Env. & horizon & AURCC$(\downarrow)$ & RPP$(\downarrow)$ & $CR_K(\uparrow)$ &  loss$(\downarrow)$ \\
        \midrule
        \multirow{4}{3.6em}{large}  & 20.0 & $0.028\pm(<0.001)$ & $0.021\pm(<0.001)$ & $0.62\pm0.035$ & $0.059\pm(<0.001)$ \\
        & 30.0 & $0.118\pm0.013$ & $0.047\pm0.004$ & $0.8\pm(<0.001)$ & $0.295\pm0.037$ \\
        & 40.0 & $0.218\pm0.018$ & $0.087\pm0.004$ & $0.82\pm0.035$ & $0.321\pm0.027$ \\
        & 50.0 & $0.16\pm0.018$ & $0.07\pm0.005$ & $0.92\pm0.035$ & $0.247\pm0.038$ \\
        \midrule
        \multirow{4}{3.6em}{medium}  & 20.0 & $0.0\pm(<0.001)$ & $0.0\pm(<0.001)$ & $0.2\pm(<0.001)$ & $0.006\pm0.001$ \\
        & 30.0 & $0.0\pm(<0.001)$ & $0.0\pm(<0.001)$ & $0.2\pm(<0.001)$ & $0.019\pm0.001$ \\
        & 40.0 & $0.001\pm(<0.001)$ & $0.0\pm(<0.001)$ & $0.2\pm(<0.001)$ & $0.032\pm0.003$ \\
        & 50.0 & $0.001\pm(<0.001)$ & $0.001\pm(<0.001)$ & $0.2\pm(<0.001)$ & $0.031\pm(<0.001)$ \\
        \midrule
        \multirow{4}{3.6em}{open}  & 20.0 & $0.005\pm(<0.001)$ & $0.001\pm(<0.001)$ & $0.2\pm(<0.001)$ & $0.029\pm0.001$ \\
        & 30.0 & $0.015\pm0.002$ & $0.007\pm0.001$ & $0.5\pm(<0.001)$ & $0.104\pm0.006$ \\
        & 40.0 & $0.036\pm0.002$ & $0.016\pm0.001$ & $0.6\pm(<0.001)$ & $0.119\pm0.007$ \\
        & 50.0 & $0.062\pm0.002$ & $0.025\pm0.001$ & $0.6\pm(<0.001)$ & $0.148\pm0.006$ \\
        \midrule
        \multirow{4}{3.6em}{umaze}  & 20.0 & $0.0\pm(<0.001)$ & $0.0\pm(<0.001)$ & $0.2\pm(<0.001)$ & $0.007\pm0.001$ \\
        & 30.0 & $0.0\pm(<0.001)$ & $0.0\pm(<0.001)$ & $0.2\pm(<0.001)$ & $0.012\pm0.002$ \\
        & 40.0 & $0.006\pm0.001$ & $0.002\pm(<0.001)$ & $0.2\pm(<0.001)$ & $0.048\pm0.003$ \\
        & 50.0 & $0.035\pm0.002$ & $0.008\pm0.001$ & $0.4\pm(<0.001)$ & $0.195\pm0.007$ \\
        \bottomrule
    \end{tabular}
\end{sc}
\end{footnotesize} 
\end{center} 
\end{table}

\begin{figure}[h!]
\centering
\includegraphics[scale=0.65]{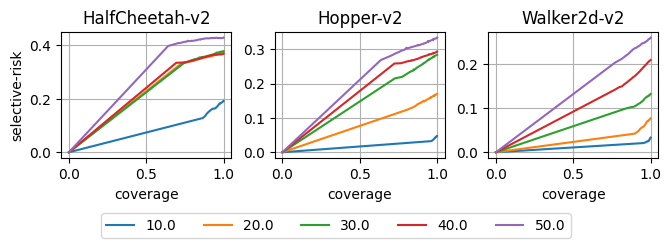}
\caption{Selective-risk coverage curves for  \emph{horizon} in \emph{gym-mujoco} environments. } 
\label{fig:gym-mujoco-horizon}
\end{figure}
\begin{figure}[h!]
\centering
\includegraphics[scale=0.65]{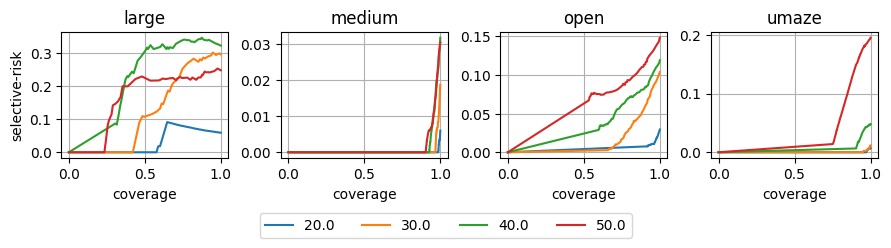}
\caption{Selective-risk coverage curves for  \emph{horizon} in \emph{maze} environments } 
\label{fig:maze-horizon}
\end{figure}

\newpage
\subsection{Ensemble-Count}
\begin{table}[h!]
\caption{Evaluation metrics for  \emph{ensemble-count} comparison in \emph{maze} environments.} 
\label{table:maze-ensemble-count}
\begin{center} 
\begin{footnotesize} 
\begin{sc}
    \begin{tabular}{|l | c | c | c| c | c | c |}
        \toprule
        Env. & ensemble-count & AURCC$(\downarrow)$ & RPP$(\downarrow)$ & $CR_K(\uparrow)$ &  loss$(\downarrow)$ \\
        \midrule
        \multirow{5}{3.6em}{large}  & 10 & $0.168\pm0.044$ & $0.051\pm0.011$ & $0.6\pm(<0.001)$ & $0.307\pm0.061$ \\
        & 20 & $0.149\pm0.039$ & $0.057\pm0.015$ & $0.78\pm0.066$ & $0.269\pm0.065$ \\
        & 40 & $0.138\pm0.031$ & $0.056\pm0.011$ & $0.82\pm0.035$ & $0.264\pm0.044$ \\
        & 80 & $0.146\pm0.019$ & $0.061\pm0.007$ & $0.82\pm0.035$ & $0.269\pm0.027$ \\
        & 100 & $0.14\pm0.015$ & $0.062\pm0.004$ & $0.82\pm0.035$ & $0.251\pm0.029$ \\
        \midrule
        \multirow{5}{3.6em}{medium}  & 10 & $0.001\pm(<0.001)$ & $0.0\pm(<0.001)$ & $0.2\pm(<0.001)$ & $0.023\pm0.007$ \\
        & 20 & $0.001\pm(<0.001)$ & $0.0\pm(<0.001)$ & $0.2\pm(<0.001)$ & $0.024\pm0.006$ \\
        & 40 & $0.001\pm(<0.001)$ & $0.0\pm(<0.001)$ & $0.2\pm(<0.001)$ & $0.021\pm0.003$ \\
        & 80 & $0.001\pm(<0.001)$ & $0.0\pm(<0.001)$ & $0.2\pm(<0.001)$ & $0.022\pm0.001$ \\
        & 100 & $0.001\pm(<0.001)$ & $0.0\pm(<0.001)$ & $0.2\pm(<0.001)$ & $0.022\pm0.001$ \\
        \midrule
        \multirow{5}{3.6em}{open}  & 10 & $0.033\pm0.004$ & $0.009\pm(<0.001)$ & $0.48\pm0.035$ & $0.127\pm0.015$ \\
        & 20 & $0.031\pm0.003$ & $0.01\pm(<0.001)$ & $0.5\pm(<0.001)$ & $0.117\pm0.019$ \\
        & 40 & $0.03\pm0.003$ & $0.011\pm0.001$ & $0.5\pm(<0.001)$ & $0.11\pm0.014$ \\
        & 80 & $0.029\pm0.002$ & $0.012\pm0.001$ & $0.5\pm(<0.001)$ & $0.108\pm0.007$ \\
        & 100 & $0.029\pm0.001$ & $0.012\pm0.001$ & $0.5\pm(<0.001)$ & $0.107\pm0.005$ \\
        \midrule
        \multirow{5}{3.6em}{umaze}  & 10 & $0.011\pm0.002$ & $0.002\pm(<0.001)$ & $0.3\pm(<0.001)$ & $0.073\pm0.006$ \\
        & 20 & $0.009\pm0.001$ & $0.002\pm(<0.001)$ & $0.3\pm(<0.001)$ & $0.074\pm0.004$ \\
        & 40 & $0.008\pm0.001$ & $0.002\pm(<0.001)$ & $0.3\pm(<0.001)$ & $0.077\pm0.004$ \\
        & 80 & $0.008\pm0.001$ & $0.002\pm(<0.001)$ & $0.3\pm(<0.001)$ & $0.075\pm0.004$ \\
        & 100 & $0.008\pm0.001$ & $0.002\pm(<0.001)$ & $0.3\pm(<0.001)$ & $0.075\pm0.003$ \\
        \bottomrule
    \end{tabular}
\end{sc}
\end{footnotesize} 
\end{center} 
\end{table}

\begin{figure}[h!]
\centering
\includegraphics[scale=0.7]{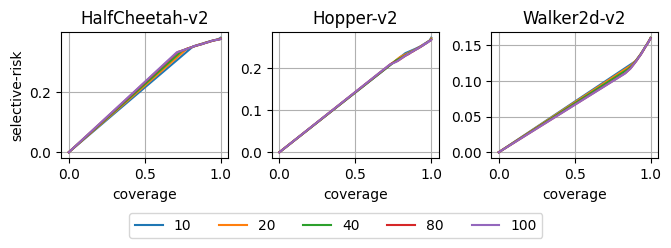}
\caption{Selective-risk coverage curves for  \emph{ensemble-count} in \emph{gym-mujoco} environments } 
\label{fig:gym-mujoco-ensemble-count}
\end{figure}
\begin{figure}[h!]
\centering
\includegraphics[scale=0.7]{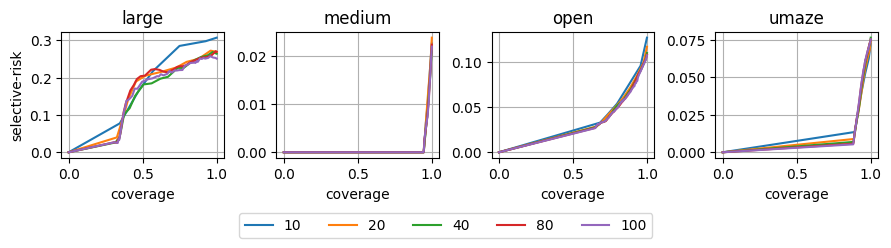}
\caption{Selective-risk coverage curves for  \emph{ensemble-count} in \emph{maze} environments } 
\label{fig:maze-ensemble-count}
\end{figure}

\end{document}